\documentclass[final,1p,times,preprint]{elsarticle}
\usepackage{amssymb}
\usepackage{amsmath}
\usepackage{lineno}

\usepackage[ruled,vlined]{algorithm2e}
\usepackage{bbding}
\usepackage{graphicx}
\usepackage{multirow}
\usepackage{booktabs}
\usepackage{colortbl}
\usepackage{xcolor}
\usepackage{url}
\usepackage{subcaption}
\usepackage{xspace}
\usepackage{tikz}
\usepackage{svg}
\usetikzlibrary{positioning, arrows.meta}
\usepackage{float}
\usepackage{soul}

\definecolor{skyblue}{RGB}{135, 206, 235}
\definecolor{peachpuff}{RGB}{255, 218, 185}

\begin{document}

\begin{frontmatter}



\title{Neural Network Surrogate and Projected Gradient Descent for Fast and Reliable Finite Element Model Calibration: a Case Study on an Intervertebral Disc}



\author[label1,label2]{Matan Atad\corref{cor1}}
\ead{matan.atad@tum.de}
\author[label1]{Gabriel Gruber}
\author[label3,label4]{Marx Ribeiro}
\author[label5]{Luis Fernando Nicolini}
\author[label1,label2]{Robert Graf}
\author[label1,label2]{Hendrik Möller}
\author[label1,label6]{Kati Nispel}
\author[label2]{Ivan Ezhov}
\author[label2,label7]{Daniel Rueckert}
\author[label1]{Jan S. Kirschke}
\cortext[cor1]{Corresponding author}

%% Author affiliations
\affiliation[label1]{organization={Institute for Neuroradiology, TUM University Hospital, School of Medicine and Health, Technical University of Munich (TUM)},city={Munich}, country={Germany}}
            
\affiliation[label2]{organization={Chair for AI in Healthcare and Medicine, Technical University of Munich (TUM) and TUM University Hospital},city={Munich}, country={Germany}}

\affiliation[label3]{organization={Department of Trauma and Reconstructive Surgery, University Hospital Halle, Martin-Luther-University Halle-Wittenberg}, city={Halle (Saale)},
            country={Germany}}

\affiliation[label4]{organization={Department of Mechanical Engineering, Federal University of Santa Catarina},
            city={Florianópolis},
            country={Brazil}}

\affiliation[label5]{organization={Department of Mechanical Engineering, Federal University of Santa Maria},
            city={Av. Santa Maria},
            country={Brazil}}

\affiliation[label6]{organization={Associate
Professorship of Sport Equipment and Sport Materials, School of Engineering and Design, Technical
University of Munich},
            city={Munich},
            country={Germany}}

\affiliation[label7]{organization={Department of Computing, Imperial College London},
           city={London},
           country={United Kingdom}}
            
%% Abstract
\begin{abstract}

Accurate calibration of finite element (FE) models is essential across various biomechanical applications, including human intervertebral discs (IVDs), to ensure their reliability and use in diagnosing and planning treatments. However, traditional calibration methods are computationally intensive, requiring iterative, derivative-free optimization algorithms that often take days to converge.

This study addresses these challenges by introducing a novel, efficient, and effective calibration method demonstrated on a human L4-L5 IVD FE model as a case study using a neural network (NN) surrogate. The NN surrogate predicts simulation outcomes with high accuracy, outperforming other machine learning models, and significantly reduces the computational cost associated with traditional FE simulations. Next, a Projected Gradient Descent (PGD) approach guided by gradients of the NN surrogate is proposed to efficiently calibrate FE models. Our method explicitly enforces feasibility with a projection step, thus maintaining material bounds throughout the optimization process.

The proposed method is evaluated against state-of-the-art Genetic Algorithm (GA) and inverse model baselines on synthetic and \textit{in vitro} experimental datasets. Our approach demonstrates superior performance on synthetic data, achieving a Mean Absolute Error (MAE) of 0.06 compared to the baselines' MAE of 0.18 and 0.54, respectively. On experimental specimens, our method outperforms the baseline in 5 out of 6 cases. While our approach requires initial dataset generation and surrogate training, these steps are performed only once, and the actual calibration takes under three seconds. In contrast, traditional calibration time scales linearly with the number of specimens, taking up to 8 days in the worst-case. Such efficiency paves the way for applying more complex FE models, potentially extending beyond IVDs, and enabling accurate patient-specific simulations.

\end{abstract}

%% Keywords

\end{frontmatter}

%\linenumbers


\section{Introduction}

Finite element (FE) simulations are a well-established tool across various fields in biomechanics and are integral to applications in orthopedics, tissue modeling, and spinal research \cite{marinkovic2019survey,naoum2021finite,basaran2019finite,haddas2019finite}.
However, accurate numerical modeling encounters substantial challenges, such as structural complexity, diverse material behaviors, and patient-specific variability \cite{karajan2012multiphasic,dreischarf2014comparison}.

One of the most demanding aspects of FE modeling is \textit{Calibration} or \textit{Parameter Estimation}, which ensures model accuracy by adjusting input material parameters until the model output closely matches \textit{in vitro} experimental or \textit{in vivo} patient-specific measurements \cite{schmidt2006application}. This process is crucial because relying solely on literature-based material properties often fails to capture the specific mechanical behaviors of tissues, such as intervertebral discs (IVDs), under varying geometries or defect conditions \cite{damm2020lumbar,schmidt2007application}. In IVD modeling, for instance, calibration is necessary to overcome the high variability and uncertainty in material parameter definitions across studies, which arises from differences in measurement methods and specimen characteristics \cite{nicolini2022experimental, schlager2018uncertainty,wang2021prediction}. This challenge is not limited to IVD models, as uncertainties in material parameters and model complexity affect the accuracy of FE simulations in various biomechanical applications \cite{wille2016uncertainty,merema2021patient}. Therefore, calibration is essential to make FE models more reliable for clinical application.

In traditional calibration approaches, researchers adjust material parameters within estimated physiological ranges to achieve the best fit between simulation and experimental outcomes \cite{schmidt2007application,schmidt2006application,heuer2007stepwise,zhang2021finite}.
This calibration process can be viewed as a constrained optimization problem that, due to the resource and time demands of a single FE simulation, can take several hours to multiple days to complete \cite{alizadeh2020managing}. FE models typically lack derivatives because they solve complex partial differential equations by discretizing the model into finite elements, leading to approximate numerical solutions without explicit derivatives \cite{li2023finite}. Consequently, derivative-free optimization algorithms, such as evolutionary algorithms, are commonly employed to accelerate the calibration \cite{nicolini2022experimental,ezquerro2011calibration,gruber12comparative}. These algorithms treat the FE model as a black box, iteratively adjusting its inputs based on the simulation outcomes. Nevertheless, calibration remains time-consuming; in a recent study by \citet{gruber12comparative}, more than 17 hours were needed to calibrate an FE model of the human IVD using a single \textit{in vitro} specimen as reference. 

A trend in biomechanical modeling is using Machine Learning (ML) surrogate models to predict numerical simulation results even when trained on a relatively small number of examples \cite{kudela2022recent,phellan2021real}. Once validated to achieve a sufficient level of approximation, they can be used as an alternative to the costly FE computations, providing near-instant predictions. Various surrogate models have been utilized \cite{hammer2024new,Ge2023,cai2021surrogate}, with a notable recent increase in the use of neural networks (NNs) \cite{milicevic2022huxley,lostado2017improvement,dalton2023physics,hsu2011comparison,lee2018optimization,sajjadinia2022multi}, due to their ability to capture complex, nonlinear relationships in data. When it comes to calibration, most surrogate-based works still use gradient-free optimization with genetic algorithms (GAs) \cite{moeini2023surrogate,lostado2017improvement,lee2018optimization,da2021optimization}, replacing the costly FE simulation with the surrogate's efficient prediction. Others suggest training a separate ML model in the inverse setting, directly predicting FE model input parameters from its outputs \cite{liu2021computation,babaei2022machine}.

Gradients derived from NNs have been effectively employed in optimization problems across various domains, such as improving model robustness \cite{goodfellow2015explaining} and explaining its predictions \cite{mothilal2020explaining}. 
Inspired by these approaches, we leverage gradients from the surrogate model to guide the calibration process, thereby improving the efficiency of parameter estimation. In biomechanics, NN surrogate gradients have been used to calibrate a left ventricle FE model, incorporating an auxiliary optimization objective that implicitly supports feasibility \cite{maso2021efficient}. In contrast, our approach explicitly enforces feasibility, setting it apart from this prior work.

In this study, we focus on the L4-L5 IVD as a case study due to its clinical relevance, nonlinear behavior, significant motion, and exposure to high loads \cite{nicolini2022effects,boden1990abnormal,ruberte2009influence,elfering2002young}. This IVD segment is often affected by degenerative diseases, making it a common target for surgical interventions and biomechanical research.

The objectives of this work are two-fold. First, we implement and evaluate the performance of multiple ML models as surrogates for an established L4-L5 IVD FE model \cite{gruber12comparative, nicolini2022experimental}. To our knowledge, we are the first to develop an NN surrogate for an IVD FE model. Second, we introduce a novel, efficient, and effective calibration method guided by the NN surrogate using Projected Gradient Descent (PGD). 
Unlike other methods that rely on independent optimization frameworks, such as mesh-based solvers or GAs, treating the surrogate as a black box \cite{moeini2023surrogate,lostado2017improvement,lee2018optimization,da2021optimization,gruber12comparative}, our approach directly leverages gradients from the surrogate, enabling efficient calibration while explicitly ensuring solution feasibility \cite{maso2021efficient}.
We compare our calibration method to a state-of-the-art GA \cite{gruber12comparative} and a developed inverse model \cite{liu2021computation,babaei2022machine} by evaluating it on a synthetic dataset and real-world experimental measurements, highlighting the practical applicability and limitations of the proposed approach. Our code is available at \url{https://github.com/matanat/IVD-CalibNN/}.

\section{Materials and methods} \label{sec2:methods}

\begin{figure}[t!]
    \centering
    \includegraphics[width=\textwidth]{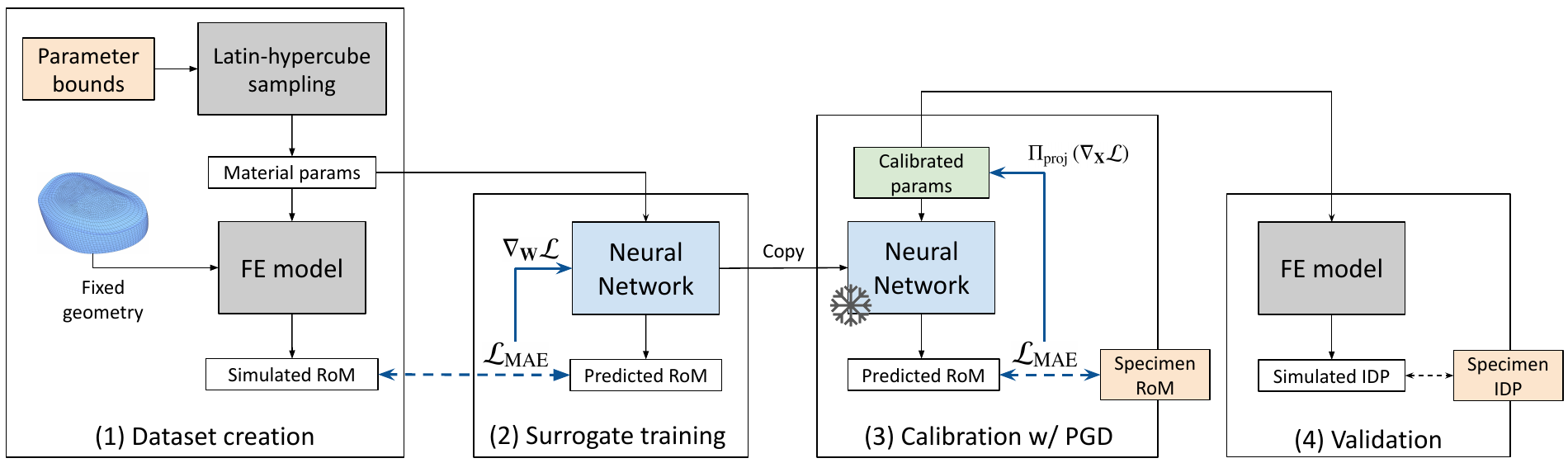}
\caption{\textbf{Calibrating an L4-L5 intervertebral disc (IVD) finite element (FE) model to match \textit{in vitro} measurements.} (1) Create a dataset by sampling material parameters within feasible bounds using Latin hypercube sampling (LHS) \cite{mckay2000comparison} and obtaining corresponding range of motion (RoM) values with FE simulations. (2) Train a neural network (NN) surrogate to minimize the Mean Absolute Error (MAE) between the predicted and simulated RoM. (3) Freeze the network weights and optimize the NN input parameters to match the predicted RoM to the measurements. Projected Gradient Descent (PGD) ensures the optimized parameters remain within feasible bounds. Finally, (4) validate the calibrated material parameters by comparing simulated intradiscal pressure (IDP) to experimental measurements.}
    \label{fig:graphical_abstract}
\end{figure}

We present a method for calibrating an FE model to match experimental measurements and demonstrate its applicability on an L4-L5 IVD FE (Fig.~\ref{fig:graphical_abstract}). First, we create a dataset using Latin hypercube sampling (LHS) \cite{mckay2000comparison} and generate corresponding FE simulation range of motion (RoM) outputs. Next, we train a surrogate NN model to approximate the FE model's behavior, enabling efficient predictions. We calibrate the FE model to experimental measurements by optimizing the input parameters with PGD. During this calibration step, the weights of the pre-trained NN model are frozen, ensuring that only the inputs are adjusted and leveraging the surrogate's learned representation. Finally, we validate the calibrated parameters by comparing their respective simulated intradiscal pressure (IDP) to experimental measurements.

We begin this section by describing the FE model and the dataset used for training the surrogate models. Afterward, we detail the surrogate model architectures and the training process. Finally, we explain the calibration process, including the optimization method, the baseline comparisons to a GA and an inverse model and the validation process.

\subsection{FE model}\label{sec:fe_model}

\begin{figure}[t!]
    \centering
    \resizebox{\textwidth}{!}{%
    \begin{tikzpicture}
        % figure
        \node[anchor=north west] (img) at (0,0) {\includegraphics[width=6cm, trim=220 0 260 0, clip]{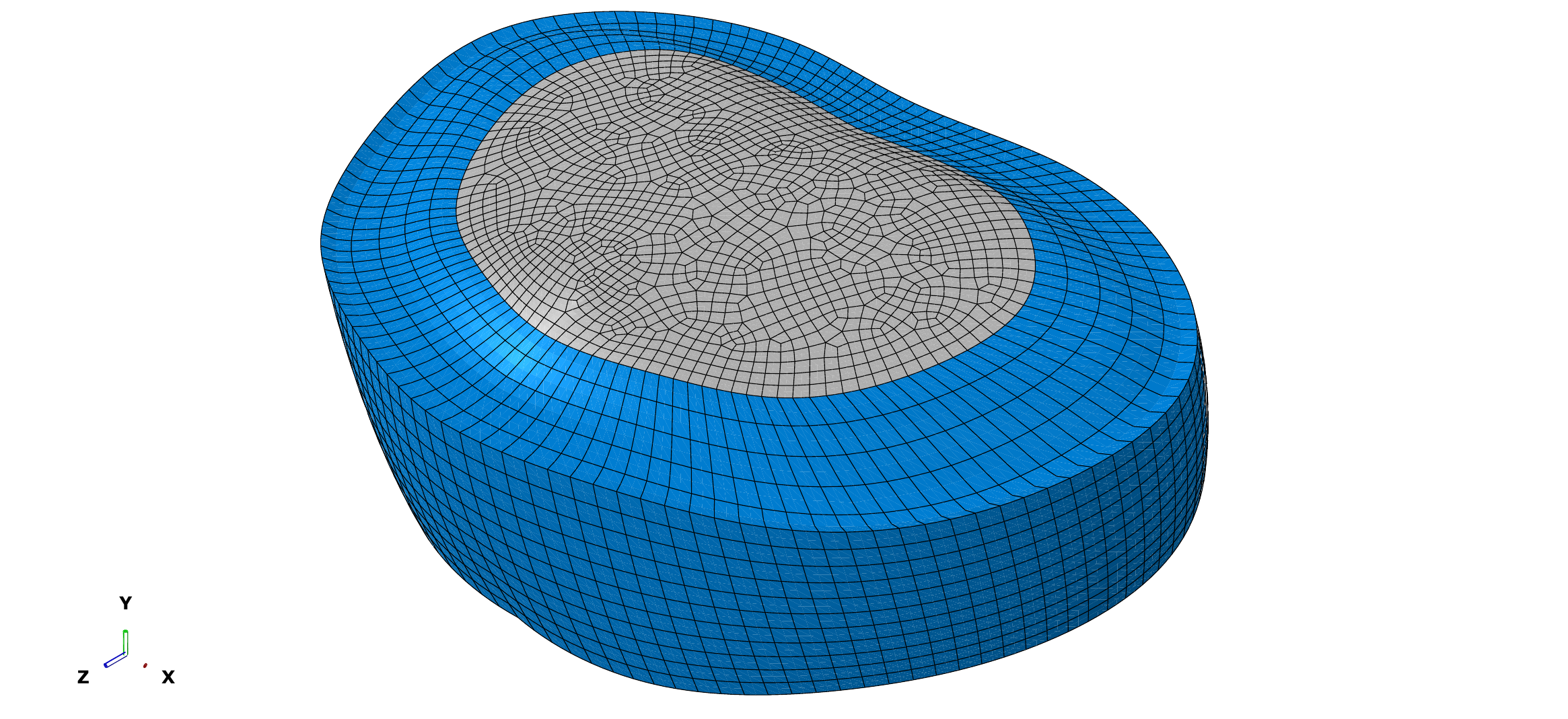}};
        % equations
        \node[anchor=west] (eq1) at (5,0) 
        {$
        \mathit{W_{a}} = \mathbf{C_{10a}}(I_{1} - 3) + \frac{1}{D}\left(\frac{J^2 - 1}{2} - \ln(J)\right) + \frac{\mathbf{k_{1}}}{2\mathbf{k_{2}}} \sum \left\{\exp\left[\mathbf{k_{2}}\langle E \rangle ^2 \right] - 1\right\}
        $};
        \node[anchor=west, below=of eq1, yshift=0.5cm] (eq2) 
        {$
        \mathit{W_{n}} = \mathbf{C_{10n}}(I_{1} - 3) + \mathbf{C_{01n}}(I_{2} - 3) + D(J - 1)^2
        $};
        % arrows
        \draw[-{Circle[scale=0.6, fill=black]}, very thick] (eq1.west) -- ([yshift=1.9cm,xshift=0.3cm]img.center);
        \draw[-{Circle[scale=0.6, fill=black]}, very thick] (eq2.west) -- ([yshift=0.7cm]img.center);
    \end{tikzpicture}
    }
    \caption{\textbf{Meshed geometry of the L4-L5 IVD and the strain energy density functions in the FE model \cite{nicolini2022experimental, gruber12comparative}}. The figure shows the separation into nucleus pulposus (gray) and annulus fibrosus (blue), both with hyperelastic material definitions. The strain energy density functions for the nucleus ($\mathit{W_{n}}$, Mooney-Rivlin) and annulus ($\mathit{W_{a}}$, Holzapfel-Gasser-Ogden) are provided, with the latter specifically accounting for the anisotropic properties resulting from the collagen fibers in the annulus. The calibration process involved 13 parameters, with five in the strain energy density functions (bold). The remaining parameters, used in functions not provided here, describe fiber dispersion, angle variations, and stiffness changes across the annulus. For a more detailed explanation, refer to our previous publications \cite{nicolini2022experimental, gruber12comparative}.}
    \label{fig:fe_model}
\end{figure}
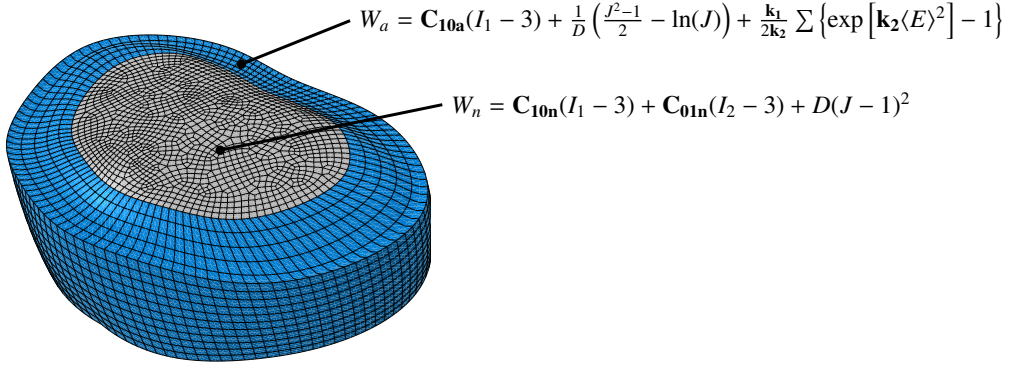

We used an established FE model of the human L4-L5 IVD \cite{nicolini2022experimental,gruber12comparative}.
The IVD geometry used in the FE model (Fig.~\ref{fig:fe_model}) was manually reconstructed based on average dimensions reported in the literature. The model separates the nucleus pulposus (gray) and the annulus fibrosus (blue), both characterized by hyperelastic material properties, with the nucleus pulposus comprising 44\% of the total disc volume. The annulus fibrosus was segmented into five symmetrical subregions, each containing five layers to reflect radial differences in the material properties of the IVD. 
The Mooney-Rivlin material model ($\mathit{W_{n}}$) described the material behavior of the nucleus pulposus, while the Holzapfel-Gasser-Ogden material model ($\mathit{W_{a}}$) \cite{holzapfel2005single} was used for the annulus fibrosus to account for its anisotropic properties resulting from its collagen fibers \cite{schmidt2006application}. These strain energy density functions, $\mathit{W_{n}}$ and $\mathit{W_{a}}$, describe the energy stored in a material per unit volume during deformation \cite{yamashita2023calculation}. The IVD was simulated without adjacent vertebrae, with a reference point 10 mm above the center of the disc's cross-sectional surface for load application \cite{nicolini2022experimental}. A coupling constraint connected this point to the upper surface of the IVD, while the lower surface was fixed via a coupling constraint to an anchored reference point, replicating the fixed caudal vertebra in \textit{in vitro} experiments. The model was meshed using quadratic hexahedral elements. 

The modeling process, combined with the specific material parameters from the strain energy density functions, resulted in 13 input parameters characterizing the behavior of the FE model. 

\subsection{Dataset creation}\label{sec:dataset}

As a preprocessing step, we considered a global sensitivity analysis of the same FE model, performed in a parallel work \cite{Gruber2024Sensitivity}. This analysis revealed that some input parameters had significant influence while others had minimal effect. However, the influence of most parameters varied across load scenarios. To avoid missing important interactions or non-linear effects, we decided to include all 13 parameters in the surrogate model training setting.

Let \( \mathbf{p}_i = [p_1, \ldots, p_d] \in \mathbb{R}^d \) represent a set of material configurations, where \( d = 13 \) denotes the dimensionality of the parameter space and \( 1 \leq i \leq n \) indexes the individual configurations. The configurations were sampled with LHS, thus ensuring a comprehensive exploration of the parameter ranges by dividing them into equally probable intervals. Each input parameter \( p_j \) was restricted to a feasible range as detailed in \ref{appendix:matrial_params}. Subsequently, we experimented with surrogates trained on configuration sets of various sizes.

The sampled configurations $\mathbf{p}$, each comprising d parameters, were used to perform simulations with the FE model. Besides $\mathbf{p}$, the simulation inputs a \textit{Load Case} definition specifying the direction and a \textit{Moment} magnitude of the applied mechanical loading. Specifically, load cases \(c \in (1, 2, 3, 4)\) were used, representing axial rotation, extension, flexion, and lateral bending, as detailed by \citet{heuer2007stepwise}. For each load case, five different moment magnitudes \( m \in (1, 2, 3, 4, 5) \), measured in Nm, were applied. The output of each simulation was the rotational displacement of the IVD, expressed as the RoM in degrees \( r_{c,m} \in \mathbb{R}_{> 0}\). This RoM represents the rotational change around the axis corresponding to the applied load case, measured relative to the initial position.

The simulations were performed on an AMD Ryzen 7 7700X machine with an NVIDIA GeForce RTX 4090 GPU, averaging 208 seconds per simulation. Input configurations for which the FE simulation did not converge were removed from the dataset, and both input and output values were normalized to the range (0,1) with min-max normalization based on the defined parameter ranges. 

\subsection{Surrogate models}

A surrogate ML model $\mathcal{M}$ was trained to predict a single RoM value $r_{c,m} \in \mathbb{R}^{0}$ for given material parameters $\mathbf{p} \in \mathbb{R}^d$ conditioned on a load case $c \in \mathbb{R}$ and a moment $m \in \mathbb{R}$. Formally, this is represented:
\begin{align}
    \label{eq:surrogate}
    \mathcal{M}: [c, m, \mathbf{p}] \rightarrow r_{c,m}
\end{align}
Predicting a single RoM value for each load case and moment combination allows the model to more accurately capture the dependencies between material parameters and RoM.

Following the literature \cite{kudela2022recent}, we evaluated Linear Regression (LR), Polynomial Regression (PR), Support Vector Machine for Regression (SVR) \cite{brereton2010support}, Random Forest (RF), Light Gradient-Boosting Machine (LightGBM) \cite{ke2017lightgbm}, Gaussian Process (GP), and NN surrogates. We used off-the-shelf implementations from scikit-learn \cite{pedregosa2011scikit} for LR, PR, SVR, RF, and GP, and the official one for LightGBM \cite{ke2017lightgbm}. The NN was implemented with PyTorch \cite{paszke2019pytorch}. Hyperparameters for all models were searched with Optuna \cite{akiba2019optuna}, with details provided in~\ref{appendix:hyperparams}. The models were trained and evaluated on the same machine mentioned before.

\paragraph{Neural network (NN)} \label{sec:nn_model}

We employed a simple feed-forward architecture for the NN consisting of five linear layers, each followed by a ReLU activation function. The final layer outputs the predicted RoM value and is not constrained to a specific range (e.g., (0,1)), allowing for later extrapolation to unseen RoM values during calibration for some specimens. The loss function was the Mean Absolute Error (MAE) between the predicted and ground-truth RoM values.

The network weights were optimized using the Adam optimizer. An early stopping mechanism was incorporated to halt training if no improvement in the loss was observed over the validation set, preventing overfitting. We applied L2 regularization to the network's weights to further mitigate overfitting and used Dropout \cite{srivastava2014dropout}, where randomly selected NN weights were zeroed during training. Dropout encourages the network to learn robust features and reduces dependency on any single neuron, enhancing generalization.

\subsection{FE model calibration}

The calibration process determines input parameters that match reported \textit{in vitro} RoM measurements by minimizing the difference between the surrogate outputs and the measurements. The developed method utilized an optimization stirred with gradients from the NN surrogate model. We compared it to a GA \cite{gruber12comparative}, which used the NN's output instead of the FE model's (GA w/ NN) and a developed inverse model.

\subsubsection{Projected Gradient Descent with NN (PGD w/ NN)}

The pre-trained NN surrogate guides the search for input parameters $\mathbf{x}_{\text{cal}} \in \mathbb{R}^{d}$ to achieve desired RoM values $\mathbf{y} \in \mathbb{R}^{k}$ for all load cases and moments specified by an experimental measurement, i.e., $k = |LoadCases| \times |Moments|$. We employed Projected Gradient Descent (PGD) \cite[p.~263]{bubeck2015convex} to optimize $\mathbf{x}_{\text{cal}}$ so that the network output matches $\mathbf{y}$. During this process, the NN weights were frozen and \textit{not changed}, ensuring that the model's learned representation remained consistent and that only the input parameters were adjusted. This process differs from the one used during the training phase of the NN, where the weights were updated to learn the mapping from inputs to outputs.

PGD is an optimization technique explicitly maintaining the feasibility of the solution in settings where traditional gradient descent may violate physical or practical constraints. It incorporates into gradient descent's update step a projection function to ensure that the updated parameters remain within predefined constraints:
\begin{align}
    \label{eq:pgd}
    \mathbf{X}^{(t+1)} = \Pi_{\text{proj}} \left(\mathbf{X}^{(t)} - \eta \nabla_{\mathbf{X}^{(t)}} \mathcal{L} \right)
\end{align}
where $\eta$ is the learning rate, $\nabla_{\mathbf{X}} \mathcal{L}$ is the gradient of the loss function with respect to $\mathbf{X}$ and $\Pi_{\text{proj}}$ is a projection function (as described below). This explicit enforcement of constraints distinguishes our method from the approach by \citet{maso2021efficient}, which used a penalty loss to implicitly encourage feasible solutions.

\begin{algorithm}[t!]
\caption{PGD w/ NN calibration}
\label{alg:ngps}
\SetAlgoLined
\KwIn{\\
\quad $\mathbf{X}_{\text{search}} \in \mathbb{R}^{k \times (2 + d)}$: Search matrix.\\
\quad $T$: Number of optimization steps.\\
\quad $\mathbf{NN}_{\text{\SnowflakeChevron}}$: Pretrained NN  with frozen weights.\\
\quad $\mathbf{y} \in \mathbb{R}^{k}$: Target specimen RoM column.\\
\quad $\eta$: Learning rate.\\
}
\KwOut{\\
\quad $\mathbf{x}_{\text{cal}} \in \mathbb{R}^{d}$: Calibrated  parameters}
\BlankLine
\BlankLine
$\mathbf{X}^{(1)} \leftarrow \mathbf{X}_{\text{search}}$; \hfill \tcp{Initialize \(\mathbf{X}\)}
\For{$t = 1$ \KwTo $T$}{
    $\hat{\mathbf{y}} \leftarrow \mathbf{NN}_{\text{\SnowflakeChevron}} \left( \mathbf{X}^{(t)} \right)$; \hfill \tcp{NN inference}
    $\mathcal{L}_{\text{MAE}} \leftarrow \frac{1}{k} \sum_{i=1}^{k} \| \mathbf{y}_i - \hat{\mathbf{y}}_i \|_1 $; \hfill \tcp{MAE loss}
    $\mathbf{X}^{(t+1)} \leftarrow {\Pi}_{\text{proj}} \left(\mathbf{X}^{(t)} - \eta \cdot  \nabla_{\mathbf{X}^{(t)}} \mathcal{L} \right)$; \hfill \tcp{Project optimized parameters}
}
$\mathbf{x}_{\text{cal}} \leftarrow \mathbf{X}^{(T)}_{0,}$; \hfill \tcp{Extract parameter values}
\BlankLine
\Return $\mathbf{x}_{\text{cal}}$
\end{algorithm}

Alg.~\ref{alg:ngps} provides an overview of the calibration process. It inputs the target RoM measurements $\mathbf{y} \in \mathbb{R}^{k}$ and a search matrix $\mathbf{X}_{\text{search}} \in \mathbb{R}^{k \times (2 + d)}$, in which the first two columns are set to combinations of load cases $\mathbf{c} \in \mathbb{R}^{c}$ and moments $\mathbf{m} \in \mathbb{R}^{m}$, specifying the conditions for which the target RoM values $\mathbf{y} \in \mathbb{R}^{k}$ are given. The other $d$ columns are set with some random input vector $\textbf{x}_{init} \sim U (0, 1)^d$.

In a loop, the NN infers RoM values $\hat{\mathbf{y}}$ for the current input, and an MAE loss for the ground-truth $\mathbf{y}$ is computed. Next, the gradient of the loss function with respect to the input $\nabla_{\mathbf{X}} \mathcal{L}$ is derived, and a PGD step is performed to update $\mathbf{X}$. These steps are performed only on the optimized parameters while leaving the first two columns of $\mathbf{X}$ intact to not change the condition load cases and moments. The process is repeated for $T$ steps until it converges. Finally, the calibrated parameter values $\mathbf{x}_{\text{cal}}$ are extracted from one of the rows of $\mathbf{X}$ (as all the rows are identical, details to follow) and returned.

\begin{figure}[tbp]
    \centering
    \includegraphics[width=0.8\textwidth]{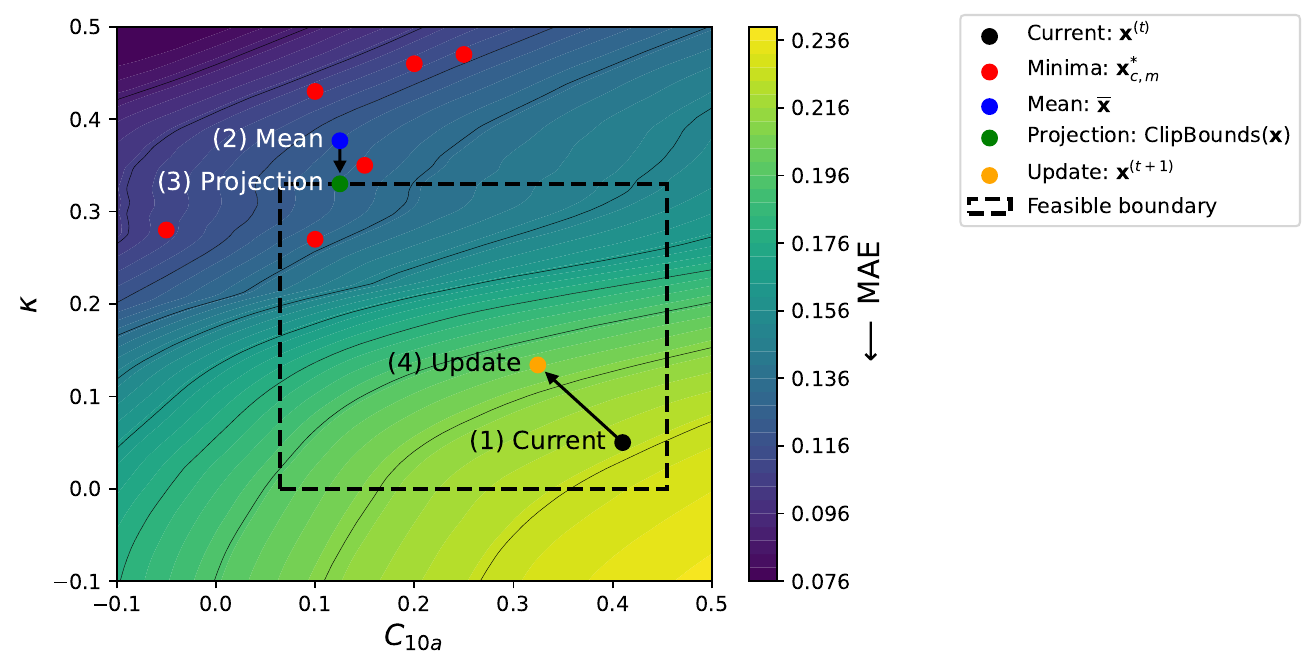}
    \caption{\textbf{Illustration of Projected Gradient Descent (PGD) in the loss landscape of two calibrated parameters.} The contour plot shows the RoM MAE between the surrogate NN's predictions and the third specimen by \citet{nicolini2022effects}. Only two input parameters are varied for illustration, while the others are fixed. The dashed box represents the feasible input parameter area. The algorithm follows these steps: (1) It starts from the black point, indicating the current step's configuration. (2) Using gradients of the loss, the red points representing local minima for specific load case and moment combinations are obtained, and their mean is computed, shown by the blue point. (3) This mean is then projected into the feasible area, shown by the green point. (4) Finally, the algorithm updates the configuration toward the projection, leading to the orange point, with the step size depending on the learning rate.}
    \label{fig:pgd_update_step}
\end{figure}

The projection function operates as follows:
\begin{align}
\label{eq:proj_1}
    \Pi_{\text{proj}} \left( \mathbf{X} \right) &= \mathbf{1}_k \cdot  \text{ClipBounds}(\overline{\mathbf{x}})\\
\label{eq:proj_2}
    \text{where} \quad \overline{\mathbf{x}} &= \frac{1}{k} \sum_{k=1}^{k} \mathbf{X}_{k,} , \quad \overline{\mathbf{x}} \in \mathbb{R}^{d} ,\\
    \text{ClipBounds} (\mathbf{x}) &= \min(\max(\mathbf{x}, \mathbf{b}_{\text{min}}), \mathbf{b}_{\text{max}})
\label{eq:proj_3}
\end{align}
Initially (Eq.~\ref{eq:proj_2}), it calculates the mean value across the rows of $\mathbf{X}$, thus ensuring a single candidate value per input parameter across all load cases and moments. Then (Eq.~\ref{eq:proj_3}), it enforces predefined constraints by clipping the values within the bounds $\mathbf{b}_{\text{min}}, \mathbf{b}_{\text{max}} \in \mathbb{R}^d$. These bounds are the same as those used for sampling the training dataset in Sec.~\ref{sec:dataset}. Finally (Eq.~\ref{eq:proj_1}), it inflates the result back to a matrix form by multiplying it from the left with a vector of $k$ ones $\mathbf{1}_k$, thus creating a matrix of $k$ identical rows. Fig.~\ref{fig:pgd_update_step} further illustrates this projection process.

In practice, Alg.~\ref{alg:ngps} was run $M=500$ times with different random initialization of $\mathbf{x}_{\text{init}}$ set into $\mathbf{X}_{\text{search}}$ and the best solution in terms of $\overline{\mathcal{R}^2}$ score\footnote{See definition of this score in Sec.~\ref{sec:metrics}.} across the different load cases was chosen. The multiple runs did not affect the overall search time, as the algorithm was implemented to perform all optimization rounds in parallel. The hyperparameters $\eta$, $T$ and $M$ were chosen using Optuna \cite{akiba2019optuna}. We used the Adam optimizer in Eq.~\ref{eq:pgd} instead of a vanilla Gradient Descent for improved results.

\subsubsection{Calibration baselines}

\paragraph{GA w/ NN}

We employed a GA \cite{nicolini2022experimental,gruber12comparative} as a first baseline, substituting the FE model in the original works with the NN surrogate. We opted for this approach because running the calibration without the surrogate would be infeasible due to the excessive running time. We used the settings from the original works \cite{nicolini2022experimental,gruber12comparative} while restricting the sampled parameters bounds as described previously and increasing the $\overline{\mathcal{R}^2}$ stopping criterion to 1. We increased this criterion for fairer comparison with the developed method and since using the surrogate circumvents previous computational restrictions. Additional details are provided in~\ref{appendix:baselines}.

\paragraph{Inverse model}

Following the approach of \cite{liu2021computation,babaei2022machine}, we developed a model to predict material parameters from an input vector of RoM values. The training setup is thus inverse to the one used for the surrogates:
\begin{align}
\mathcal{M}_{Inverse}: [r_{1,1}, .., r_{1,5}, .., r_{4,1}, .., r_{4,5}] \rightarrow \mathbf{p}
\end{align}
where $\mathcal{M}_{Inverse}$ is an ML model, $\mathbf{p} \in \mathbb{R}^d$, and $r_{c,m} \in \mathbb{R}, \forall c \in (1,2,3,4)$, $\forall m \in (1,2,3,4,5)$ as defined in Sec.~\ref{sec:dataset}. Notably, $\mathcal{M}_{Inverse}$ was not conditioned on specific $c$ and $m$ as the surrogate models in Eq.~\ref{eq:surrogate}; instead, all RoM values were provided to give complete context for inferring $\mathbf{p}$.

In order to obtain a dataset for training the inverse model, we sampled random material configurations $\mathbf{p}$ within the material bounds as before and inferred their corresponding RoM values for all load cases and moments using the NN surrogate. We then trained an NN model in this inverse setting. Additional details are provided in~\ref{appendix:baselines}.

\subsection{Validation}\label{sec:validation}

Validation is performed to ensure that the computational model reflects known physiological behavior \cite{noailly2007does}. In this study, we compared simulated IDP, which was not part of the calibration process, to ranges reported in the literature \cite{brinckmann1991change}. In the simulation, IDP was measured at the center of the nucleus pulposus in megapascal (MPa), with observed values ranging between 0 and 2.5 MPa. In contrast to the load cases used for training the surrogate, validation was performed using axial compression loading with an applied compressive force ranging from 0 to 2000 Newtons (N).

\subsection{Evaluation metrics}\label{sec:metrics}

We evaluated the surrogate models and the calibration methods using the Mean Absolute Error (MAE), commonly used to evaluate regression problems. The MAE between the ground truths RoM $y$ and the outputs $\hat{y}$ is averaged over load cases and moments as follows:
\begin{align}
    \text{MAE} &= \frac{\sum_{c \in \text{LoadCases}}{\sum_{m \in \text{Moments}}{| y_{c,m} - \hat{y}_{c,m} |}}}{|\text{LoadCases}| \times |\text{Moments}|}
\end{align}
For the surrogate models, the ground truth $y$ is the FE numerical simulation result, and the output $\hat{y}$ is the surrogate predicted value. For the calibration methods, the ground truth $y$ is the experimental measurement, and the output $\hat{y}$ is the FE numerical result for the calibrated configuration. 

Additionally, we used the mean $\mathcal{R}^2$-score, denoted $\overline{\mathcal{R}^2}$, which measures the proportion of variance explained across different load cases \citep{wirthl2023global, schmidt2006application}:
\begin{align}
    \overline{\mathcal{R}^2} &= \frac{\sum_{c \in \text{LoadCases}}{\mathcal{R}^2_c}}{|\text{LoadCases}|}\\
    \text{where} \ \forall c \in \text{LoadCases}, \quad \mathcal{R}^2_c &= 1 - \frac{\sum_{m \in \text{Moments}}{\left( y_{c,m} - \hat{y}_{c,m} \right)^2}}{\sum_{m \in \text{Moments}}{\left( y_{c,m} - \overline{y_c} \right)^2}}
\end{align}
where $\overline{y_c}$ is the mean of the ground truths across moments for load case $c$.\footnote{\citet{wirthl2023global} use a different term for an identically computed metric. For clarity, we adopt the more widely accepted $\mathcal{R}^2$ for both.}

\section{Results and discussion} \label{sec:discussion}

This section compares the performance of the different surrogate models used to approximate FE model simulations. We discuss the results of the calibration methods by evaluating them on both synthetic and experimental data. Then, we present the results of the validation process. Finally, we conduct ablation studies to understand the impact of various components of our approach. Detailed results are provided in~\ref{appendix:detailed_results}.

\subsection{Surrogate models}

To enhance the efficiency and accuracy of FE calibration, we explore using surrogate models as a faster and more reliable alternative to direct simulations. This section assesses the performance of various surrogate models, focusing on the effects of model architecture and training dataset size. Additionally, it investigates the models’ ability to interpolate and extrapolate RoM values to unseen moments, which is essential for calibrating to experimental data. The following experiments were designed to identify the best-performing surrogate model with rigorous evaluation on large test sets, where the predicted RoM values are compared to ones obtained from FE simulations. In the subsequent sections, the best-performing surrogate was evaluated on its ability to calibrate material parameters with experimental data, where only a limited number of samples are available.

\subsubsection{Model architecture} 

To assess the performance of various surrogate model architectures, we trained the models on a dataset of 1024 LHS samples and evaluated them on a test set of 64 randomly sampled configurations. All experiments were conducted with 4-fold cross-validation.

In the last two columns of Table~\ref{tab:dataset_size}, models commonly used in Response Surface methods \cite{chakraborty2014adaptive,hammer2024new}, such as LR and PR, result in poor performance. In particular, the PR models show improvement with increasing polynomial degrees of 4 and 5; however, they remain prone to overfitting, as indicated by their high MAE (0.49 and 0.53, respectively). More advanced methods, such as GP and LightGBM, show stronger performance (MAE of 0.33 and 0.28). Nevertheless, the NN outperforms all others, with an MAE of 0.10 and $\overline{\mathcal{R}^2}$ of 0.99, highlighting its superior ability to handle the non-linear interactions in the FE simulations \cite{nicolini2022experimental}. As surrogate accuracy is crucial for the calibration task, the NN was chosen for the subsequent experiments.

\begin{table}[t!]
\centering
\resizebox{\textwidth}{!}{%
\begin{tabular}{llcccccccccc}
\toprule
\multicolumn{2}{l}{\multirow{2}{*}{Surrogate model}} & \multicolumn{2}{c}{$n=128$} & \multicolumn{2}{c}{$n=512$} & \multicolumn{2}{c}{$n=1024$} \\
\cmidrule(r){3-4} \cmidrule(r){5-6} \cmidrule(r){7-8}
 & & $\overline{\mathcal{R}^2}$ $\uparrow$ & $\text{MAE}$ $\downarrow$ & $\overline{\mathcal{R}^2}$ $\uparrow$ & $\text{MAE}$ $\downarrow$ & $\overline{\mathcal{R}^2}$ $\uparrow$ & $\text{MAE}$ $\downarrow$ \\
\midrule
\multicolumn{2}{l}{Linear Regression (LR)} & 0.55$\pm$0.13 & 1.03$\pm$0.29 & 0.54$\pm$0.15 & 1.05$\pm$0.26 & 0.55$\pm$0.15 & 1.05$\pm$0.26 \\
\multicolumn{2}{l}{Support Vector Regression (SVR)} & 0.59$\pm$0.11 & 1.05$\pm$0.15 & 0.65$\pm$0.13 & 0.98$\pm$0.14 & 0.51$\pm$0.25 & 1.04$\pm$0.06 \\
\multirow{4}{*}{Polynomial Regression (PR),} & degree 2 & 0.62$\pm$0.07 & 0.96$\pm$0.23 & 0.68$\pm$0.08 & 0.85$\pm$0.21 & 0.70$\pm$0.09 & 0.81$\pm$0.22\\
& degree 3 & 0.62$\pm$0.05 & 0.94$\pm$0.19 & 0.79$\pm$0.07 & 0.68$\pm$0.19 & 0.82$\pm$0.07 & 0.61$\pm$0.22\\
 & degree 4 & 0.53$\pm$0.16 & 0.99$\pm$0.11 & 0.77$\pm$0.09 & 0.70$\pm$0.10 & 0.88$\pm$0.03 & 0.49$\pm$0.11\\
 & degree 5 & 0.45$\pm$0.21 & 1.07$\pm$0.11 & 0.74$\pm$0.10 & 0.75$\pm$0.10 & 0.86$\pm$0.03 & 0.53$\pm$0.10\\
\multicolumn{2}{l}{Random Forest (RF)} & 0.82$\pm$0.03 & 0.65$\pm$0.17 & 0.89$\pm$0.06 & 0.46$\pm$0.16 & 0.91$\pm$0.04 & 0.40$\pm$0.12 \\
\multicolumn{2}{l}{Gaussian Process (GP)} & 0.83$\pm$0.04 & 0.62$\pm$0.18 & 0.92$\pm$0.01 & 0.43$\pm$0.09 & 0.95$\pm$0.01 & 0.33$\pm$0.09 \\
\multicolumn{2}{l}{LightGBM} & 0.85$\pm$0.01 & 0.57$\pm$0.12 & 0.95$\pm$0.02 & 0.34$\pm$0.09 & 0.96$\pm$0.01 & 0.28$\pm$0.07 \\
\multicolumn{2}{l}{Neural Network (NN)} & \textbf{0.90$\pm$0.02} & \textbf{0.43$\pm$0.12} & \textbf{0.98$\pm$0.00} & \textbf{0.16$\pm$0.04} & \textbf{0.99$\pm$0.00} & \textbf{0.10$\pm$0.03} \\
\midrule
\multicolumn{2}{l}{Dataset creation time (days)} & \multicolumn{2}{c}{1.2} & \multicolumn{2}{c}{4.9} & \multicolumn{2}{c}{9.8} \\
\bottomrule
\end{tabular}
}
\caption{Mean $\overline{\mathcal{R}^2}$ (higher is better) and MAE (lower is better) scores for surrogates trained on different dataset sizes, with standard deviations on cross-validated folds. Also shown is the time required to create each dataset.}
\label{tab:dataset_size}
\end{table}

\subsubsection{Training set size}

The size of the dataset used for training surrogate models is critical due to the time-consuming nature of FEM simulations. Table~\ref{tab:dataset_size} explores the number of samples required for accurate model predictions, comparing datasets of 128, 512, and 1024 samples. Increasing the dataset size significantly improves most models. For instance, the RF model's MAE decreases from 0.65 to 0.40 as the dataset grows from 128 to 1024, demonstrating that larger datasets help models capture underlying patterns more effectively. 
Notably, the NN model's MAE of 0.43 for 128 samples may already be sufficient for some applications, requiring only about 1.2 days to create the dataset. Observing that the NN surrogate begins to converge at 512 samples, we selected the best-performing NN model, trained on 1024 samples. 

\subsubsection{Interpolation and extrapolation} \label{sec:intra_extra}

\begin{figure}[t!]
    \centering
    \includegraphics[width=0.85\textwidth]{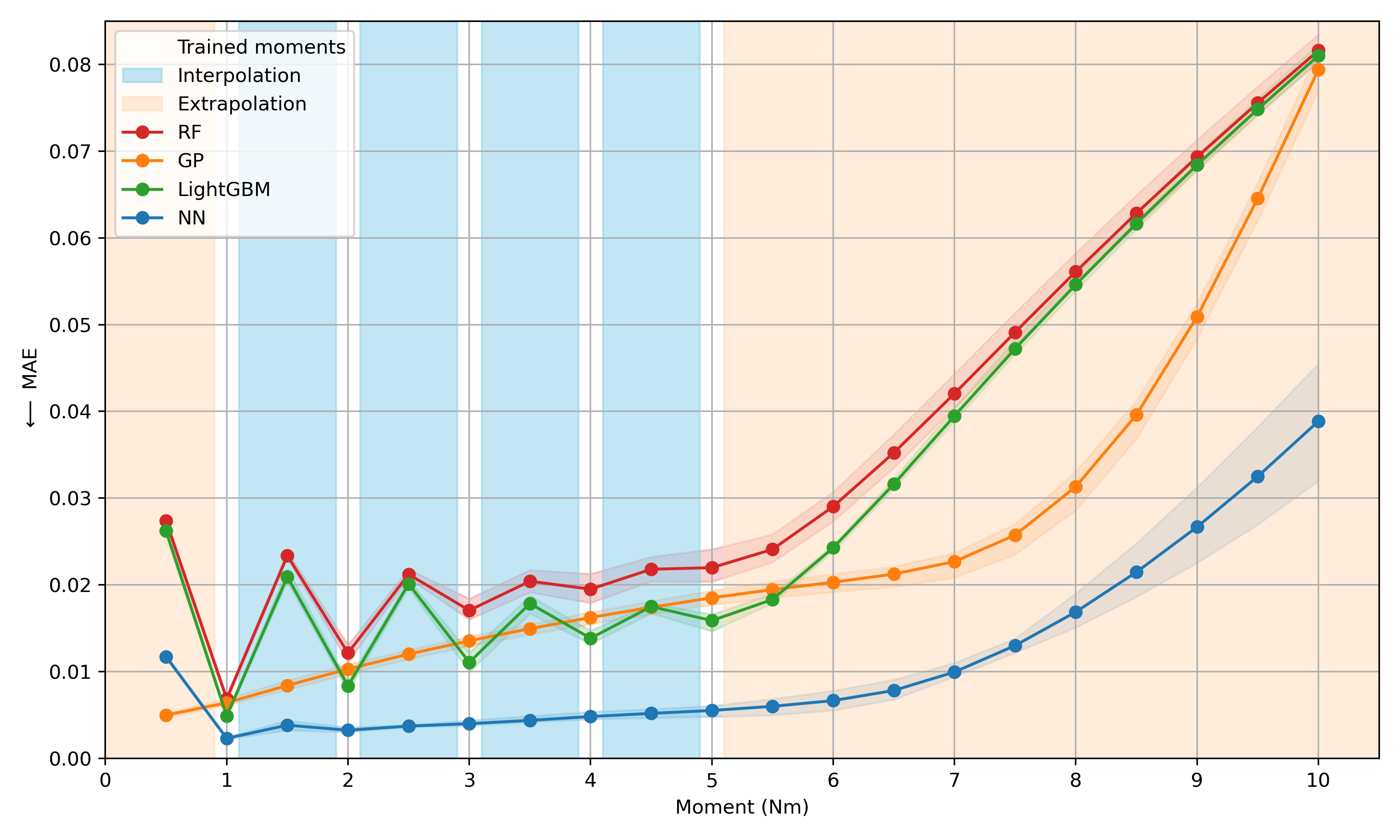}
    \caption{\textbf{Interpolation and extrapolation abilities of surrogate models across moments}. The background colors correspond to the ranges evaluated: white for trained moment, peach for extrapolation, and blue for interpolation. Both abilities are essential for the calibration of specimens. The line shadings represent the standard deviation across the cross-validated folds. The NN surrogate outperforms the other models for all but moments of 0.5 Nm.}
    \label{fig:interextra}
\end{figure}

The ability to interpolate and extrapolate to unseen moments is crucial for the developed calibration method, as it relies on surrogate models and experimental measurements that are often reported at moments different from those used for training the surrogate.\footnote{For instance, \citet{heuer2007stepwise} published their measurements for moments (1.25, 2.5, 3.75, 5, 6.25, 7.5, 10) Nm.} These generalization abilities are evaluated in Fig.~\ref{fig:interextra}, where the top four performing surrogate models are tested on a dataset of 64 synthetic configurations with ground-truth RoM values for moments in the range (0.5, 10) Nm at 0.5 intervals. The evaluation assesses performance on moments on which the model was trained (1, 2, 3, 4, 5), interpolated moments (1.5, 2.5, 3.5, 4.5), and extrapolated moments (0.5 and $> 5$). Cross-validated results are provided in~\ref{appendix:detailed_results}.

Decision tree-based methods, specifically RF and LightGBM, perform well on trained moments but struggle with unseen ones, indicating limited generalization ability. Interestingly, the GP model excels at a moment of 0.5 Nm with an MAE of 0.005, but this is insignificant since low-moment experimental measurements are uncommon. However, the GP also shows a significant performance drop for moments greater than 6 Nm. The NN model demonstrates superior performance across all moments above 1 Nm, achieving the lowest MAE of 0.004 on trained and interpolated moments. However, it shows increased error and variability for extremely high moments, indicating that while the NN surrogate is highly effective, it may still face challenges with extensive extrapolation.

\subsection{FE model calibration}\label{sec:caliberation_experiments}

In this section, the performance of the proposed calibration method was compared to the baselines. First, we tested the calibration methods on a synthetic dataset to establish baseline performance and computational efficiency. We then applied the calibration techniques to \textit{in vitro} experimental measurements conducted by \citet{nicolini2022effects} and \citet{heuer2007stepwise}, assessing their practical applicability and robustness.

\subsubsection{Synthetic data}

We evaluated the calibration methods using a synthetic test set of 64 sampled configurations. Each method received target RoM values for the four load cases at the five moments and was tasked with predicting the corresponding configuration parameter values. The predicted configurations were then input into the surrogate to assess the agreement between the predicted and target RoM values. Due to the considerable running time of the FE for 64 samples, we did not validate these results with simulations.

\begin{table}[t!]
\centering
\resizebox{\textwidth}{!}{%
\begin{tabular}{lcccccccccccc}
\toprule
\multirow{2}{*}{Load case} & \multicolumn{2}{c}{PGD w/ NN (Ours)} & \multicolumn{2}{c}{GA w/ NN} & \multicolumn{2}{c}{Inverse model} \\
\cmidrule(r){2-3} \cmidrule(r){4-5} \cmidrule(r){6-7}
 & $\mathcal{R}^2$ $\uparrow$ & MAE $\downarrow$ & $\mathcal{R}^2$ $\uparrow$ & MAE $\downarrow$ & $\mathcal{R}^2$ $\uparrow$ & MAE $\downarrow$ \\
\midrule
Axial Rotation & \textbf{0.99$\pm$0.01} & \textbf{0.07$\pm$0.06} & 0.93$\pm$0.08 & 0.21$\pm$0.18 & -0.04$\pm$1.57 & 0.80$\pm$0.73 \\
Extension & \textbf{0.98$\pm$0.12} & \textbf{0.07$\pm$0.18} & 0.95$\pm$0.06 & 0.24$\pm$0.20 & 0.61$\pm$0.73 & 0.59$\pm$0.46 \\
Flexion & \textbf{1.00$\pm$0.00} & \textbf{0.05$\pm$0.03} & 0.97$\pm$0.05 & 0.17$\pm$0.14 & 0.82$\pm$0.20 & 0.41$\pm$0.28 \\
Lateral Bending & \textbf{1.00$\pm$0.01} & \textbf{0.04$\pm$0.04} & 0.98$\pm$0.02 & 0.09$\pm$0.06 & 0.78$\pm$0.28 & 0.27$\pm$0.20 \\
\textit{Mean} & \textbf{0.99$\pm$0.01} & \textbf{0.06$\pm$0.02} & 0.96$\pm$0.05 & 0.18$\pm$0.13 & 0.54$\pm$0.34 & 0.52$\pm$0.20 \\
\midrule
Calibration time (seconds) & \multicolumn{2}{c}{2.58} & \multicolumn{2}{c}{11.92} & \multicolumn{2}{c}{\textbf{0.01}} \\
\bottomrule
\end{tabular}
}
\caption{Comparison of calibration methods on synthetic data, with mean and standard deviation across the test set. The evaluation is performed on the moments predicted by the surrogate for the found configurations. Calibration time on the entire test set is provided.}
\label{table:calibration_syn}
\end{table}

Table~\ref{table:calibration_syn} demonstrates that PGD w/ NN outperforms both baselines on synthetic data, achieving the lowest MAE of 0.06 and an $\overline{\mathcal{R}^2}$ score of 0.99, with a runtime of approximately 2.6 seconds. The GA w/ NN method also performs well, with an MAE of 0.18 and $\overline{\mathcal{R}^2}$ of 0.96, but it exhibits a more considerable variability across moments for most load cases. The high performance of both methods is expected since the calibrated target RoM values were derived from the same FE model used to create the training dataset. Though the inverse model is the fastest, requiring only 0.01 seconds, it fails to obtain applicable configurations, yielding a mean MAE of 0.52 and $\overline{\mathcal{R}^2}$ of 0.54. 

The inverse setting requires mapping RoM measurements back to distinct material input configurations, which is highly challenging. Many different configurations can produce nearly identical RoM values due to the non-monotonicity of the response surface \cite{Gruber2024Sensitivity}, resulting in the ill-posedness of the inverse problem \cite{o1986statistical}. Given this lack of convergence, we did not proceed with evaluating the inverse model on experimental data.

\subsubsection{Experimental specimen}

\begin{figure}[t!]
    \centering
    \begin{subfigure}[b]{\textwidth}
        \centering
        \includegraphics[width=\textwidth]{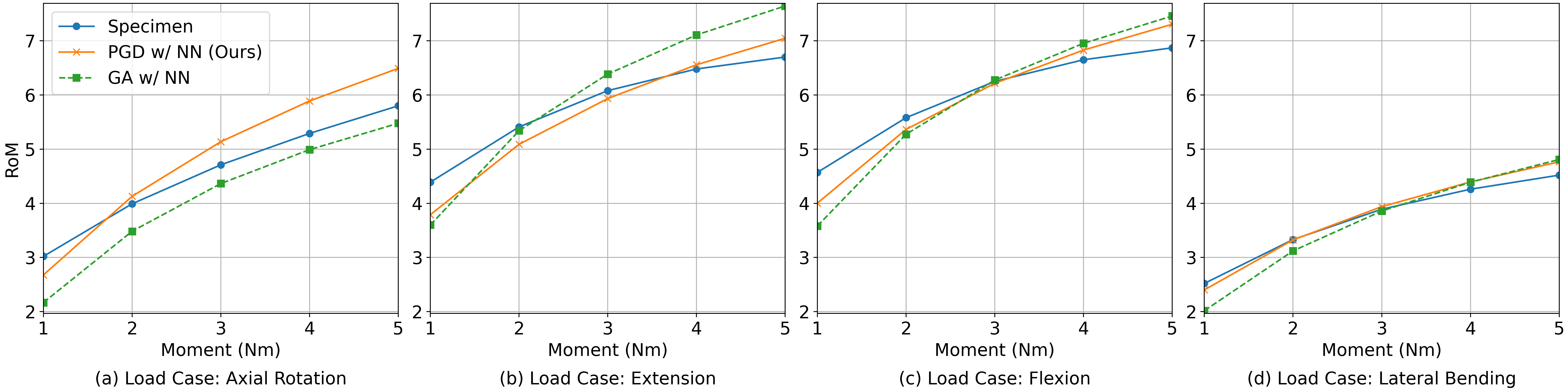}
    \end{subfigure}
    \vfill
    \begin{subfigure}[b]{\textwidth}
        \centering
        \includegraphics[width=\textwidth]{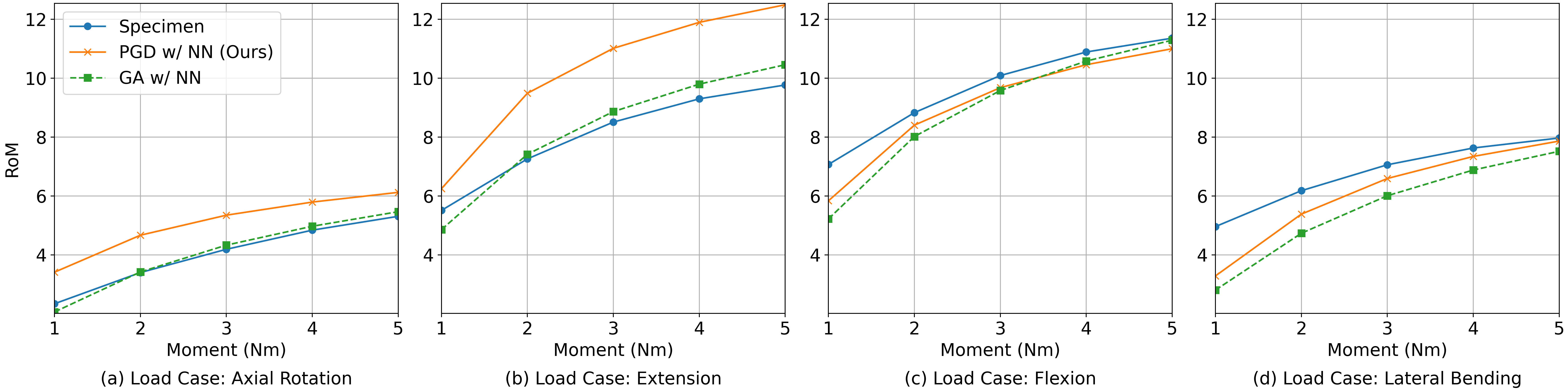}
    \end{subfigure}
    \caption{\textbf{Example calibration results for samples 1 (top) and 3 (bottom) from experiments by \citet{nicolini2022effects}.} The plots show the RoM values for the \textit{in vitro} measurements (blue), the calibrated configuration using the proposed method (orange), and the GA w/ NN baseline (green). The proposed method performs better on sample 1 than on sample 3. The poorer performance on sample 3 is attributed to the large variability in RoM across load cases in the target configuration, which differs significantly from the surrogate's training data. See~\ref{appendix:detailed_results} for the other specimens.}
    \label{fig:calibration_samples_examples}
\end{figure}

To asses their relevance to real-world problems, we evaluated the calibration methods on six \textit{in vitro} experimental specimens. We used five specimens from experiments conducted by \citet{nicolini2022effects}, reported in the same conditions used to train the surrogate. Additionally, a mean of eight specimen measurements reported by \citet{heuer2007stepwise} was utilized and preprocessed as suggested by \citet{gruber12comparative}. This measurement is available at moments different from the ones for which the surrogate was trained. Each method was given the target RoM values and tasked with predicting the corresponding configurations. Different from before, these predicted configurations were then validated with FE simulations to assess the agreement between the simulated and target RoM. Example results are plotted in Fig.~\ref{fig:calibration_samples_examples}, Table~\ref{table:experimental} provides a summary, and the calibrated configurations are given in~\ref{appendix:detailed_results}.

\begin{table}[t!]
\centering
\resizebox{0.8\textwidth}{!}{%
\begin{tabular}{llcccc}
\toprule
\multirow{2}{*}{Sample}  & & \multicolumn{2}{c}{PGD w/ NN (Ours)} & \multicolumn{2}{c}{GA w/ NN}\\
\cmidrule(r){3-4} \cmidrule(r){5-6}
 & & $\overline{\mathcal{R}^2}$ $\uparrow$ & MAE $\downarrow$ & $\overline{\mathcal{R}^2}$ $\uparrow$ & MAE $\downarrow$ \\
\midrule
\multirow{5}{*}{\citet{nicolini2022effects}} & 1 & \textbf{0.97 $\pm$ 0.02} & \textbf{0.28 $\pm$ 0.13} & 0.95 $\pm$ 0.01 & 0.42 $\pm$ 0.13 \\
 & 2 & 0.96 $\pm$ 0.07 & \textbf{0.07 $\pm$ 0.04} & \textbf{0.97 $\pm$ 0.02} & 0.10 $\pm$ 0.08 \\
 & 3 & 0.80 $\pm$ 0.17 & 1.11 $\pm$ 0.72 & \textbf{0.93 $\pm$ 0.08} & \textbf{0.62 $\pm$ 0.43} \\
 & 4 & 0.91 $\pm$ 0.07 & \textbf{0.35 $\pm$ 0.13} & \textbf{0.93 $\pm$ 0.03} & 0.37 $\pm$ 0.16 \\
 & 5 & \textbf{0.94 $\pm$ 0.08} & \textbf{0.20 $\pm$ 0.21} & 0.94 $\pm$ 0.05 & 0.27 $\pm$ 0.14 \\
\citet{heuer2007stepwise} & Mean & \textbf{0.97 $\pm$ 0.02} & \textbf{0.41 $\pm$ 0.29} & 0.95 $\pm$ 0.03 & 0.55 $\pm$ 0.31 \\
\bottomrule
\end{tabular}
}
\caption{Comparison of calibration methods on experimental specimens, showing the mean and standard deviation across load cases per sample.}
\label{table:experimental}
\end{table}

\paragraph{Optimizing for MAE} 

The proposed method shows increased performance in MAE, outperforming the baseline on five out of six specimens. However, it exceeds the baseline in $\overline{\mathcal{R}^2}$ for only three of the six samples, which aligns with the optimization’s focus on minimizing MAE. Notably, for the third sample by \citet{nicolini2022effects}, our method’s MAE is higher than the baseline’s (1.11 vs. 0.62). Both methods performed well for the mean of specimens by \citet{heuer2007stepwise} (MAE of 0.41 and 0.55, respectively), leveraging the surrogate’s interpolation and extrapolation capabilities (Sec.~\ref{sec:intra_extra}).

\paragraph{Increased error at lowest and highest moment} 

\begin{figure}[tb]
    \centering
    \includegraphics[width=0.7\textwidth]{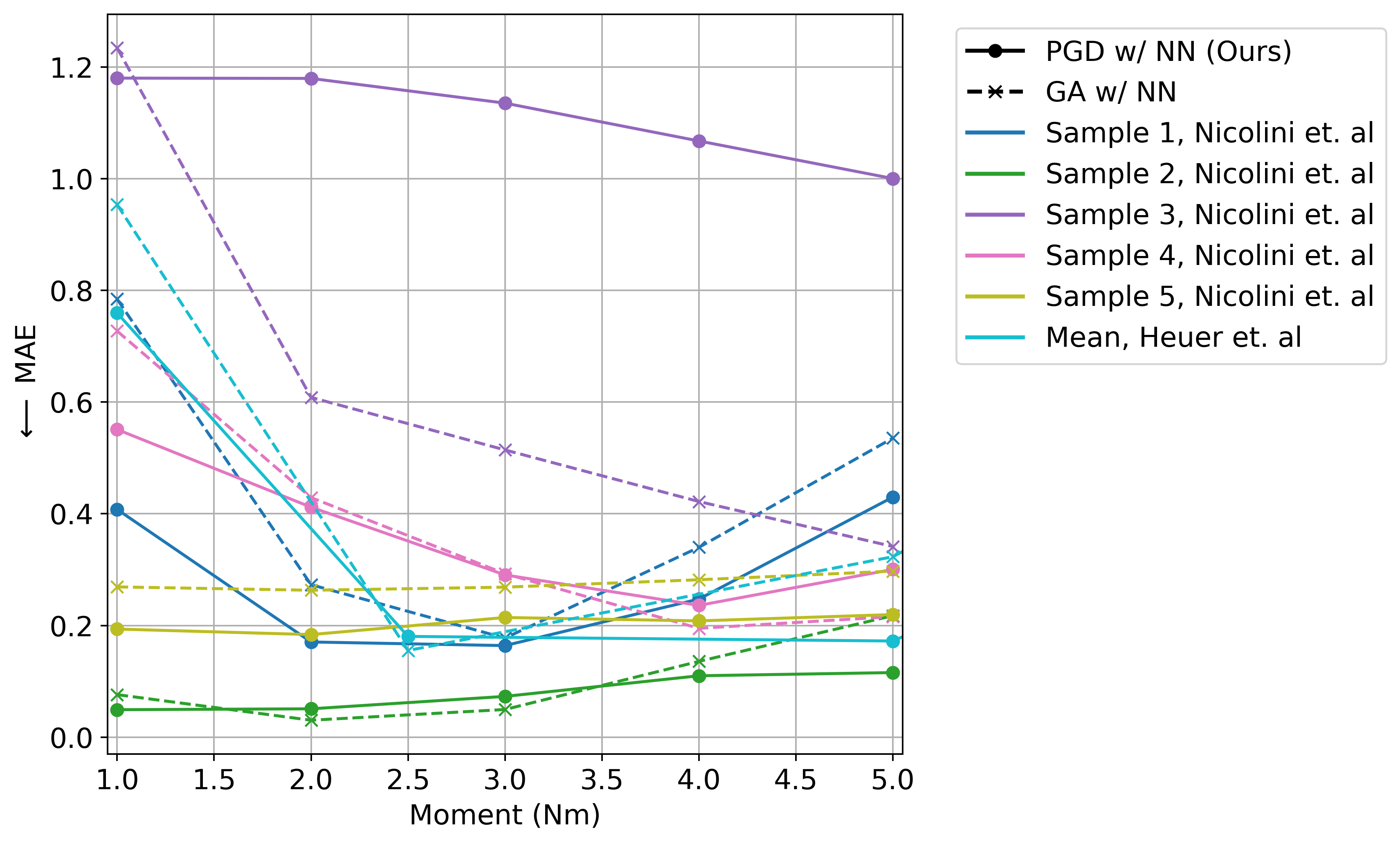}
    \caption{\textbf{Comparison of MAE vs. moment averaged across load cases on calibrated experimental specimens}. Specimens are color-coded. The proposed method outperforms the baseline except for the third sample from \citet{nicolini2022effects} (purple). Both methods present an increased error at the lowest and highest moments.}
    \label{fig:compare_calib}
\end{figure}

Fig.~\ref{fig:compare_calib} analyzes the error values across moments, averaged across load cases for experimental data. A significant increase in error is apparent at the lowest moment for many specimens. However, the surrogate evaluation in Sec.~\ref{sec:intra_extra} suggests this error does not originate from the surrogate itself. An increase in error is also noticeable at the highest moment. We hypothesize that many measurements, such as the first sample by \citet{nicolini2022effects}, exhibit high non-linearity, particularly in the transition between 1 and 2 Nm. This high non-linearity challenges the underlying FE model, producing poorer calibration results. Observations about the difficulties of the FE model in capturing diverse specimens while assuming identical geometry were reported by \citet{nicolini2022effects,nicolini2022experimental}.

\paragraph{Sensitivity to high variability in moment magnitude}

\begin{figure}[tb]
    \centering
    \begin{subfigure}[b]{0.485\textwidth}
        \centering
        \includegraphics[width=\textwidth]{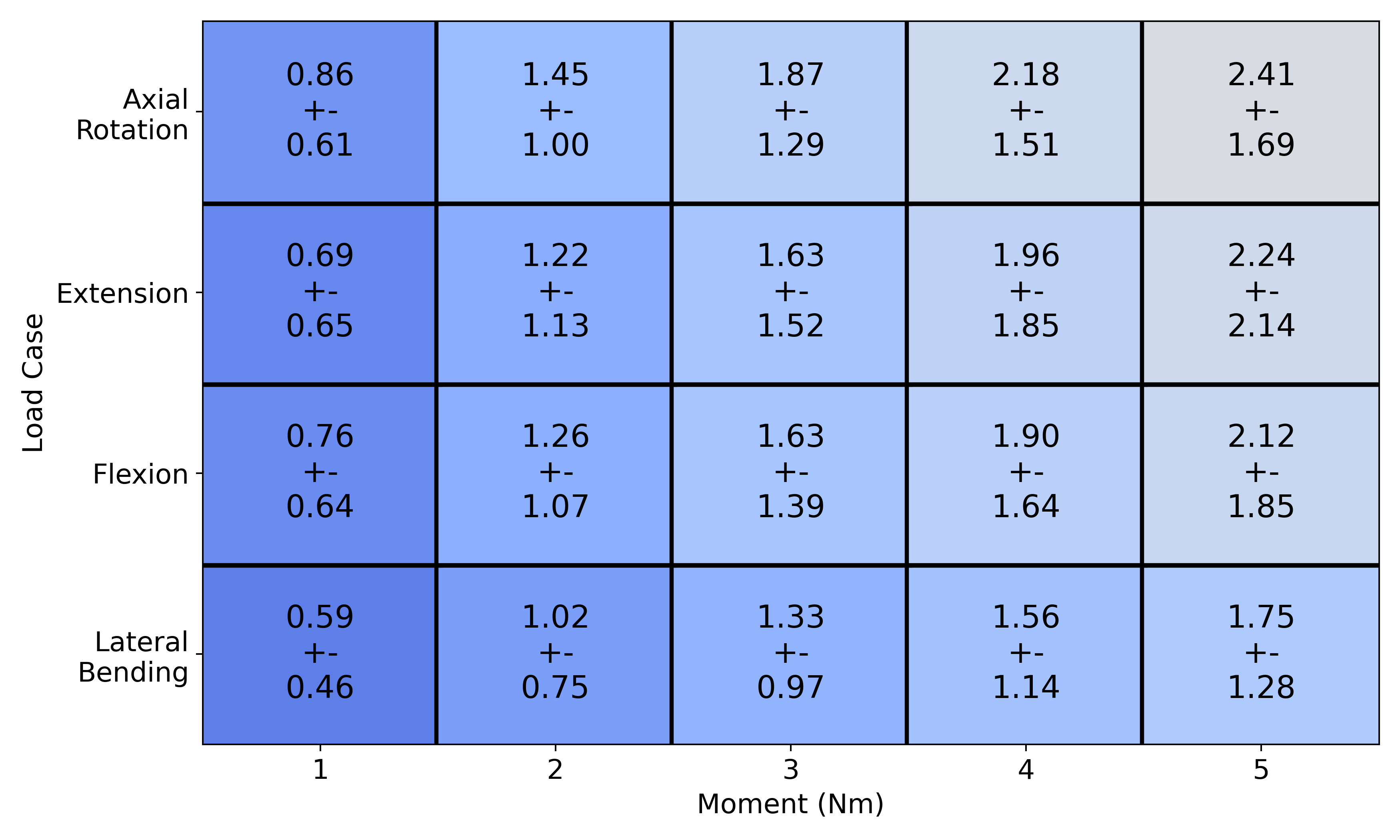}
        \caption{}
        \label{fig:rom_magnitude_diff_dataset}
    \end{subfigure}
    \begin{subfigure}[b]{0.49\textwidth}
        \centering
        \includegraphics[width=\textwidth]{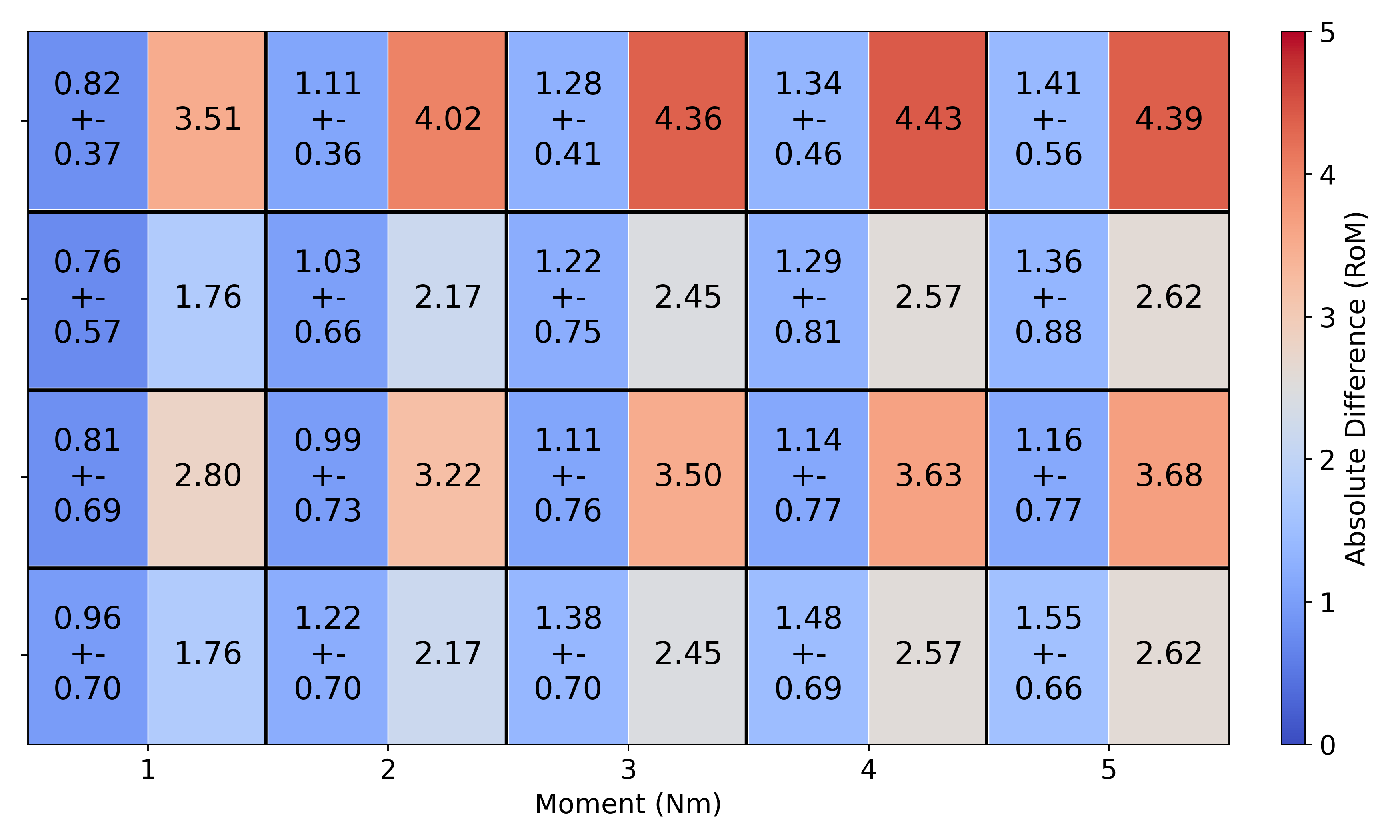}
        \caption{}
        \label{fig:rom_magnitude_diff_specimen}
    \end{subfigure}
    \caption{\textbf{Matrix comparison of absolute differences in RoM across load cases.} Each cell presents the difference between a specific RoM value and the values for the same sample at the same moment for other load cases. The left (a) shows the differences for the training set. The right (b) displays the differences for experimental values from \citet{nicolini2022effects}, with the left cell split representing all samples \textit{except} the third, and the right split showing \textit{only} the third sample. For multiple samples, the mean and standard deviation are shown. Notably, the third sample by \citet{nicolini2022effects} exhibits significantly higher differences in axial rotation and flexion.}
    \label{fig:rom_magnitude_diff}
\end{figure}

Table~\ref{table:experimental} highlights challenges with the third sample from \citet{nicolini2022effects}, which we attribute to significant variations in target RoM magnitudes. Fig.~\ref{fig:rom_magnitude_diff} presents matrices of absolute RoM differences. Each cell shows the absolute difference between a specific RoM value and the corresponding values for the same sample across the other load cases. These matrices provide a detailed comparison of how the RoM varies between different conditions for each sample. Notably, the third sample (Fig.~\ref{fig:rom_magnitude_diff_specimen}, right cell split) has much higher differences in axial rotation and flexion (4.4 and 3.7, respectively) compared to other test specimens (Fig.~\ref{fig:rom_magnitude_diff_specimen}, left cell split) and the surrogate's training set (Fig.~\ref{fig:rom_magnitude_diff_dataset}).

We hypothesize that this substantial variability in experimental data negatively impacts the optimization process. The GA w/ NN baseline, better suited for global optimization \cite{bajpai2010genetic}, excels in exploring a wider search space and finding configurations outside the training distribution. However, it still required five times more generations to converge for the third sample from \citet{nicolini2022effects} compared to the other specimens.\footnote{The GA w/ NN required 47 generations to converge to an $\overline{\mathcal{R}^2}$ higher than 0.9 for this specimen vs 9.2 generations on average for the rest.} In contrast, our method relies on NN weights derived from the training set, which do not account for such extreme configurations.

\paragraph{Computational efficiency}

We compared the efficiency of the proposed method with estimated run-times for a traditional GA that directly uses FE simulations rather than the surrogate.\footnote{Details of how we obtained these estimates are provided in~\ref{appendix:running_time}.} In Table~\ref{tab:calibration_time_comparison}, our method requires dataset generation and model training, but these are performed once, and calibration is parallelized, making its time independent of the number of specimens $n$. In contrast, while the GA does not require preliminary steps, its overall running time increases linearly with $n$. PGD w/ NN becomes more efficient than the GA already when calibrating more than one specimen in the worst case or more than three specimens in the average case.

\begin{table}[t!]
\centering
\resizebox{0.95\textwidth}{!}{%
\begin{tabular}{lcccc}
\toprule
\multirow{2}{*}{Method} & & Preliminary steps & Calibration & End-to-end \\ 
& & once & per-specimen & $n$ specimens \\
\midrule
PGD w/ NN (Ours) & & 9.8 days (dataset + training) & 2.58 sec & 9.8 days \\
\multirow{2}{*}{Traditional GA (Estimate)}  & Avg. case & \multirow{2}{*}{N/A} 
 & 2.7 days  & $2.7 \times n$ days \\
& Worst case & & 7.9 days & $7.9 \times n$ days \\
\bottomrule
\end{tabular}
}
\caption{Running time comparison between the developed method and average and worst-case estimates for a traditional GA. The running time of the developed method is independent of the number of calibrated specimens.}
\label{tab:calibration_time_comparison}
\end{table}

\subsection{Validation}

\begin{figure}[htbp]
    \centering
    \includegraphics[width=0.7\textwidth]{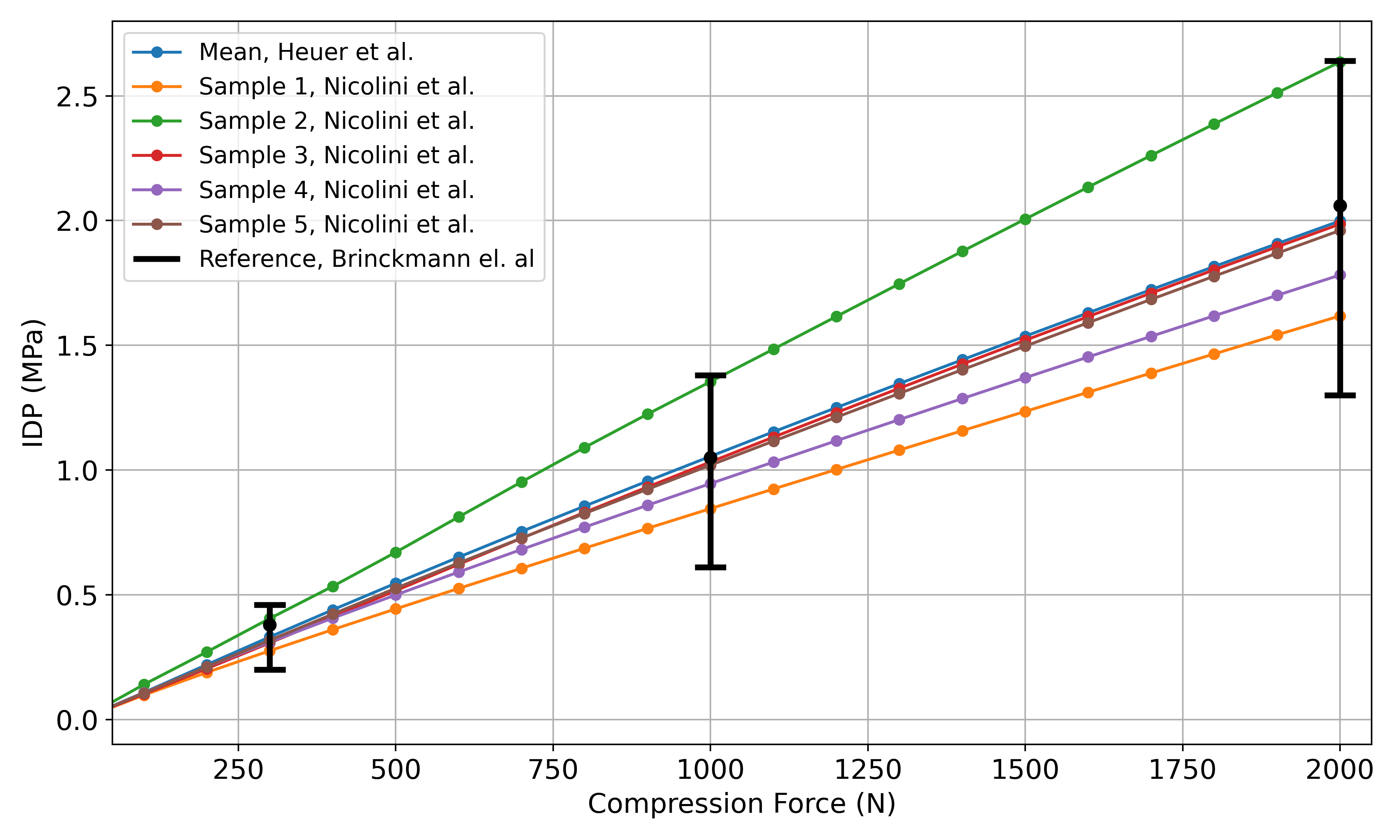}
    \caption{\textbf{Model validation of calibrated specimens with IDP under axial compression.} The black horizontal bars indicate the mean, minimum, and maximum measurements in the literature \cite{brinckmann1991change}. The colored lines show the simulation results for the calibrated specimens configurations, all within the reference ranges.}
    \label{fig:validation_IDP}
\end{figure}

To validate the calibrated configurations we compared the simulated IDP with reference ranges reported by \citet{brinckmann1991change}, who provided experimental IDP measurements under axial compression for L4-L5 IVD specimens under three applied compressive forces: 300, 1000, and 2000 N. The analysis in Fig.~\ref{fig:validation_IDP} shows that all calibrated configurations fall within the reference ranges, demonstrating that the model accurately replicates the \textit{in vitro} values observed in several human IVD specimens, ensuring the reliability of the calibration.

\subsection{Ablation studies}

To thoroughly analyze the proposed calibration method, ablation studies were performed using the experimental specimens by \citet{nicolini2022effects}. These studies focused on the impact of the projection step, loss function, and optimizer on calibration performance. Table~\ref{tab:ablation_metrics} compares the evaluated settings and includes a final column summing parameter values exceeding physical bounds, with values above zero indicating infeasible solutions. Ablation of the NN model's basic components (e.g., number of layers, activation function) was also conducted but not listed here.

The results demonstrate that the projection step is essential for ensuring feasible calibration outcomes. In the “Adam, L1 loss” setting, while the $\overline{\mathcal{R}^2}$ is highest at 0.95 and the MAE is lowest at 0.04, it results in a sum of 118.38 parameter values exceeding the bounds, making them infeasible for FE model input. With projection applied (“Adam, L1 loss, projection (Ours)”), the exceeding values drop to zero, though with a decrease in performance metrics ($\overline{\mathcal{R}^2}$ of 0.85 and MAE of 0.21).

\begin{table}[t!]
    \centering
    \resizebox{0.8\textwidth}{!}{%
    \begin{tabular}{lccc}
        \toprule
        Optimization setting & $\overline{\mathcal{R}^2}$ \ $\uparrow$ & MAE \ $\downarrow$ &
        Sum exceeding values \ $\downarrow$\\
        \midrule
        Adam, L1 loss  & \textbf{0.95} & \textbf{0.04} & 118.38 \\
        Adam, L1+bounds loss   & 0.86 & 0.21 & 0.11 \\
        Adam, L1 loss, projection (\textbf{Ours}) & 0.85 & 0.21 & \textbf{0} \\
        Adam, L2 loss, projection  & 0.84 & 0.25 & \textbf{0} \\
        L-BFGS, L1 loss, projection  & -1.0 & 0.93 & \textbf{0} \\
        \bottomrule
    \end{tabular}
    }
    \caption{Ablation studies of the proposed calibration method, evaluated on the specimens by \citet{nicolini2022effects}. The last column sums the values exceeding the material bounds, with positive values representing infeasible solutions.}
    \label{tab:ablation_metrics}
\end{table}

The “Adam, L1+bounds loss” setting, incorporating a penalty for out-of-bound values similar to \citet{maso2021efficient}, reduces infeasible values to 0.11. However, it does not eliminate them, rendering some results unusable for the FE model. Not listed here, higher penalty weights reduced exceeding values but significantly degraded performance. Using an L2 loss (“Adam, L2 loss, projection”) treats large errors more severely, encouraging the minimization of significant deviations. An NN surrogate trained with L2 loss was used for consistency in this setting. However, this loss performed slightly worse than L1 loss, with an $\overline{\mathcal{R}^2}$ of 0.84 and an MAE of 0.25. Lastly, we tried the L-BFGS optimizer instead of Adam. The “L-BFGS, L1 loss, projection” setting performed poorly, with an $\overline{\mathcal{R}^2}$ of -1.0 and an MAE of 0.93, indicating that this optimizer is unsuitable for this task. While these experiments highlight the strength of PGD in our setting, alternative methods for constrained optimization could be explored in future studies, such as Lagrange multipliers \cite{bertsekas2014constrained} or recent enhancements of PGD \cite{antonau2021relaxed}.

\section{Conclusions and outlook}\label{sec:conclusion}

This study introduced a novel calibration method guided by an NN surrogate. Using an NN surrogate provides near-instant predictions, allowing for rapid calibration through optimization with PGD. Additionally, incorporating the projection step ensures optimized parameter values remain within feasible bounds, addressing a potential pitfall in existing NN surrogate optimization-based calibration methods. This balance between performance and feasibility is crucial for practical applications in clinical settings. Moreover, the developed method outperformed the evaluated baselines. While the GA w/NN baseline was effective, it required more time and relied on an external algorithm. Despite its rapid inference, the inverse model struggled with ill-posedness, failing to produce useful material configurations. We conducted our study on an FE model of the human L4-L5 IVD, future work could extend to other biomechanical domains, as we believe our method can be adapted to other FE models that face similar challenges in calibration.

\paragraph{Surrogate modeling}

High variability in moments caused sub-optimal convergence, as the NN was not trained on such extremes. Augmenting the dataset with diverse cases could improve robustness. Future work should explore active learning \cite{echard2011ak} to generate representative samples by selectively querying the FE model for challenging, real-world scenarios. While the NN's superior accuracy is beneficial, its limited explainability remains a concern in clinical applications. Future studies should explore methods to understand its internal decision-making process \cite{reddy2022explainability}, such as the feature importance approach we used in \citet{Gruber2024Sensitivity}. Our findings there could also help reduce the surrogate's complexity by removing less impactful inputs. Finally, the GA baseline was evaluated with the NN surrogate rather than FE simulations. Despite precautions like setting a high GA stopping criterion and the NN's high accuracy, some approximation error is expected.

\paragraph{Patient-specific clinical applications}

Patient-specific models must consider disc material properties and geometric variations, especially in diseased discs with significant structural changes \cite{soltani2024ct,niemeyer2012geometry}. Medical imaging, e.g., MRI, provides a promising path to integrate specific geometry into future models \cite{strickland2011development,rayudu2022patient,nispel2024mri2fem}. Using these techniques, the surrogate model could be retrained for patient-specific geometry extracted from MRI. Another key challenge is loading estimate, which is unknown in clinical practice. Dual fluoroscopy combined with multi-body simulations could be used to better assess \textit{in vivo} loading conditions \cite{li2009segmental, li2021motion,hammer2024new,lerchl2024musculoskeletal}. Future studies could incorporate a complete functional spinal unit in the FE model, enabling the surrogate to capture IVD mechanics and interactions with surrounding ligaments and facet joints. While the NN offers precise predictions, it lacks uncertainty estimation, which could be useful for assessing unknown loading. Applying stochastic methods in future work could help address this limitation \cite{wang2021prediction}. Incorporating these factors into clinical settings will require effective calibration. Our method facilitates efficient and accurate identification of feasible parameters, representing a key step toward patient-specific simulations.

\section*{Data availability}

The data used to train the surrogate models and the synthetic data used for evaluation are available in our GitHub repository. The experimental data used to evaluate the calibration methods could be obtained by contacting the respective authors.

\section*{Acknowledgments}
Funding: This work was supported by the European Research Council (ERC) under the European Union's Horizon 2020 research and innovation program. Grant no: 101045128-iBack-epic-ERC-2021-COG

\section*{Declaration of generative AI and AI-assisted technologies in the writing process}
During the preparation of this work the the authors used ChatGPT and Grammarly to improve the readability and language of the manuscript. After using these tools, the authors reviewed and edited the content as needed and take full responsibility for the content of the published article.

\clearpage

\appendix

\section{Mathematical notation}

\begin{itemize}
    \item \textbf{Scalars, Vectors and Matrices}:
    \begin{itemize}
        \item \(x \in \mathbb{R}\): A scalar.
        \item \(\mathbf{y} \in \mathbb{R}^d\): A vector of dimension \(d\).
        \item \(\mathbf{X} \in \mathbb{R}^{n \times m}\): A matrix with \(n\) rows and \(m\) columns.
    \end{itemize}
    \item \textbf{Indices and Elements}:
    \begin{itemize}
        \item \(\mathbf{x}_i\): The \(i\)-th element of the vector \(\mathbf{x}\).
        \item \(\mathbf{X}_{i, j}\): The element in the \(i\)-th row and \(j\)-th column of the matrix \(\mathbf{X}\).
        \item \(\mathbf{X}_{i, :}, \mathbf{X}_{:, j}\): The \(i\)-th row of the matrix \(\mathbf{X}\) and the \(j\)-th column of the matrix \(\mathbf{X}\) respectively.
    \end{itemize}
    \item \textbf{Gradient and Partial Derivatives}:
    \begin{itemize}
        \item \(\nabla_{\mathbf{X}} \mathcal{L}\): The matrix of partial derivatives of the scalar loss function \(\mathcal{L}\) with respect to each element of the matrix \(\mathbf{X}\).
    \end{itemize}
\end{itemize}

\section{Detailed methodological parameters and models}
\subsection{Material parameters} \label{appendix:matrial_params}

\begin{table}[htbp]
    \centering
    \begin{tabular}{lcc}
        \toprule
        Material Parameter & Minimum Value & Maximum Value\\
        \midrule
        $C_{10n}$ & 0.03 & 0.21 \\
        $C_{01n}$ & 0.0075 & 0.0525 \\
        $C_{10a}$ & 0.065 & 0.455 \\
        $k_1$ & 1 & 50 \\
        $k_2$ & 10 & 200 \\
        $\kappa$ & 0 & 0.33 \\
        $k_{1c}$ & -0.2 & 0 \\
        $k_{2c}$ & -0.2 & 0 \\
        $k_{1r}$ & -0.2 & 0 \\
        $k_{2r}$ & -0.2 & 0 \\
        $\alpha$ & 7.5 & 52.5 \\
        $\alpha_c$ & 0 & 0.3 \\
        $\alpha_r$& 0 & 0.2 \\
        \bottomrule
    \end{tabular}
    \caption{Material parameter ranges}
    \label{tab:material_parameter_ranges}
\end{table}

\clearpage

\subsection{Calibration baselines} \label{appendix:baselines}

\paragraph{GA}

The GA \cite{nicolini2022experimental,gruber12comparative} was configured with a population size of 20, evolving from initially sampled random configurations within the parameter ranges. The six best-performing individuals from each generation are selected for \textit{Crossover} and \textit{Mutation}. During Crossover, four new individuals are created by combining pairs of the selected individuals. Mutation generates four new individuals by randomly altering a parameter of one of the selected individuals. Additionally, \textit{Immigration} is performed, introducing six new individuals per generation. Every tenth generation, a \textit{Restart} is performed and 19 new individuals are added to the best individual from the previous generation. The process iterates until an $\overline{\mathcal{R}^2}=1$ stopping criteria is met or $120$ generations are reached. 

\paragraph{Inverse model}

The inverse NN was a five-layer feed-forward network with ReLU activation functions. Output values were constrained to (0,1) using a Sigmoid function and compared to ground truths with an MAE loss. An optimizer scheduler reduced the learning rate when the loss stopped improving on the validation set, enhancing training efficiency and convergence. We trained the network on a dataset of $50,000$ samples and, as before, implemented it with PyTorch. Hyperparameters are provided in~\ref{appendix:hyperparams}.

\clearpage

\subsection{Models hyperparameters}
\label{appendix:hyperparams}

\begin{table}[htbp]
    \centering
    \resizebox{0.63\textwidth}{!}{%
    \begin{tabular}{ll}
        Model & Hyper-parameters \\
        \toprule
        LR & Default settings \\
        \midrule
        PR (degree=2) & \begin{tabular}[c]{@{}l@{}}ridge regularization,
        alpha=1.0 \end{tabular} \\
        \midrule
        PR (degree=3) & \begin{tabular}[c]{@{}l@{}}ridge regularization, 
        alpha=1.0 \end{tabular} \\
        \midrule
         PR (degree=4) & \begin{tabular}[c]{@{}l@{}}ridge regularization,
        alpha=0.11694087489403014 \end{tabular} \\
        \midrule
         PR (degree=5) & \begin{tabular}[c]{@{}l@{}}ridge regularization,
        alpha=0.7741762364391351 \end{tabular} \\
        \midrule
        RF & \begin{tabular}[c]{@{}l@{}}max\_depth=31, \\
        n\_estimators=51, \end{tabular} \\
        \midrule
        SVR & \begin{tabular}[c]{@{}l@{}}C=2.479459357922161, \\ gamma="auto", \\ kernel="rbf"\end{tabular} \\
        \midrule
        LightGBM & \begin{tabular}[c]{@{}l@{}}num\_leaves=71, \\ max\_depth=24, \\ learning\_rate=0.0959080973897064, \\ n\_estimators=183\end{tabular} \\
        \midrule
        GP & \begin{tabular}[c]{@{}l@{}}kernel=C(constant\_value=1.3535458162698054e-05) \\ * RBF(length\_scale=0.9739157609228299), \\ normalize\_y=True\end{tabular} \\
        \midrule
        NN (n=1024, n=512) & 
        \begin{tabular}[c]{@{}l@{}}batch\_size=20, \\ layers\_width=[256, 128, 64, 32, 16], \\ learning\_rate=0.000555, \\ num\_epochs=300, \\ weight\_decay=5.16293e-06, \\ early\_stopping\_patience=23 \\
        dropout\_probability=0.01
        \end{tabular} \\
        \midrule
        NN (n=128) & 
        \begin{tabular}[c]{@{}l@{}}batch\_size=14, \\ layers\_width=[128, 64, 32, 16, 8], \\ learning\_rate=0.000559, \\ num\_epochs=231, \\ weight\_decay=5.16442e-06, \\ early\_stopping\_patience=26 \\
        dropout\_probability=0.01
        \end{tabular} \\
        \midrule
        Inverse NN &
        \begin{tabular}[c]{@{}l@{}}batch\_size=48, \\ layers\_width=[128, 256, 256, 256, 128], \\ learning\_rate=0.000654, \\ num\_epochs=776, \\ weight\_decay=0.000006, \\ early\_stopping\_patience=50 \\
        dropout\_probability=0.028433,\\
        scheduler="ReduceLROnPlateau", \\
        scheduler\_patience=18, \\
        scheduler\_factor=0.05
        \end{tabular} \\
    \end{tabular}
    }
    \caption{Hyperparameters used for training the ML models}
    \label{tab:hyperparams}
\end{table}

\clearpage

\section{Detailed results}\label{appendix:detailed_results}

\begin{table}[htbp]
\centering
\resizebox{\textwidth}{!}{%
\begin{tabular}{lcccccccccc}
\toprule
\multirow{2}{*}{Model} & \multicolumn{2}{c}{Axial Rotation} & \multicolumn{2}{c}{Extension} & \multicolumn{2}{c}{Flexion} & \multicolumn{2}{c}{Lateral Bending} & \multicolumn{2}{c}{\textit{Mean}} \\
\cmidrule(r){2-3} \cmidrule(r){4-5} \cmidrule(r){6-7} \cmidrule(r){8-9} \cmidrule(r){10-11}
& $\mathcal{R}^2$ $\uparrow$ & MAE $\downarrow$ & $\mathcal{R}^2$ $\uparrow$ & MAE $\downarrow$ & $\mathcal{R}^2$ $\uparrow$ & MAE $\downarrow$ & $\mathcal{R}^2$ $\uparrow$ & MAE $\downarrow$ & $\mathcal{R}^2$ $\uparrow$ & MAE $\downarrow$ \\
\midrule
LR & 0.52$\pm$0.02 & 1.25$\pm$0.03 & 0.42$\pm$0.02 & 1.39$\pm$0.03 & 0.77$\pm$0.01 & 0.71$\pm$0.02 & 0.51$\pm$0.02 & 0.78$\pm$0.02 & 0.55$\pm$0.13 & 1.03$\pm$0.29 \\
SVR & 0.66$\pm$0.01 & 1.07$\pm$0.02 & 0.67$\pm$0.04 & 1.28$\pm$0.06 & 0.65$\pm$0.01 & 0.95$\pm$0.03 & 0.41$\pm$0.16 & 0.90$\pm$0.15 & 0.59$\pm$0.11 & 1.05$\pm$0.15 \\
PR, degree 2 & 0.54$\pm$0.03 & 1.24$\pm$0.04 & 0.60$\pm$0.01 & 1.14$\pm$0.01 & 0.74$\pm$0.02 & 0.80$\pm$0.03 & 0.62$\pm$0.05 & 0.67$\pm$0.05 & 0.62$\pm$0.07 & 0.96$\pm$0.24 \\
PR, degree 3 & 0.58$\pm$0.03 & 1.17$\pm$0.04 & 0.65$\pm$0.02 & 1.10$\pm$0.05 & 0.71$\pm$0.04 & 0.84$\pm$0.07 & 0.57$\pm$0.05 & 0.69$\pm$0.05 & 0.63$\pm$0.06 & 0.95$\pm$0.19 \\
PR, degree 4 & 0.58$\pm$0.04 & 1.10$\pm$0.04 & 0.67$\pm$0.03 & 1.12$\pm$0.09 & 0.64$\pm$0.05 & 0.89$\pm$0.07 & 0.26$\pm$0.17 & 0.88$\pm$0.11 & 0.54$\pm$0.17 & 1.00$\pm$0.11 \\
PR, degree 5 & 0.52$\pm$0.03 & 1.19$\pm$0.04 & 0.62$\pm$0.03 & 1.19$\pm$0.08 & 0.59$\pm$0.04 & 0.96$\pm$0.05 & 0.10$\pm$0.18 & 0.94$\pm$0.10 & 0.46$\pm$0.21 & 1.07$\pm$0.12 \\
RF & 0.78$\pm$0.03 & 0.80$\pm$0.04 & 0.84$\pm$0.03 & 0.78$\pm$0.05 & 0.82$\pm$0.01 & 0.65$\pm$0.01 & 0.86$\pm$0.02 & 0.36$\pm$0.02 & 0.82$\pm$0.03 & 0.65$\pm$0.17 \\
GP & 0.76$\pm$0.03 & 0.79$\pm$0.05 & 0.84$\pm$0.01 & 0.75$\pm$0.03 & 0.84$\pm$0.02 & 0.56$\pm$0.02 & 0.87$\pm$0.01 & 0.35$\pm$0.01 & 0.83$\pm$0.04 & 0.62$\pm$0.18 \\
LightGBM & 0.83$\pm$0.03 & 0.66$\pm$0.05 & 0.87$\pm$0.02 & 0.68$\pm$0.06 & 0.85$\pm$0.02 & 0.58$\pm$0.02 & 0.85$\pm$0.04 & 0.37$\pm$0.04 & 0.85$\pm$0.01 & 0.57$\pm$0.12 \\
NN & \textbf{0.86$\pm$0.02} & \textbf{0.58$\pm$0.05} & \textbf{0.91$\pm$0.02} & \textbf{0.50$\pm$0.03} & \textbf{0.91$\pm$0.01} & \textbf{0.39$\pm$0.01} & \textbf{0.93$\pm$0.02} & \textbf{0.25$\pm$0.03} & \textbf{0.90$\pm$0.02} & \textbf{0.43$\pm$0.12} \\
\bottomrule
\end{tabular}}
\caption{Surrogate models cross-validated $\mathcal{R}^2$ and MAE results. Training set $n=128$.}
\end{table}
\begin{table}[htbp]
\centering
\resizebox{\textwidth}{!}{%
\begin{tabular}{lcccccccccc}
\toprule
\multirow{2}{*}{Model} & \multicolumn{2}{c}{Axial Rotation} & \multicolumn{2}{c}{Extension} & \multicolumn{2}{c}{Flexion} & \multicolumn{2}{c}{Lateral Bending} & \multicolumn{2}{c}{\textit{Mean}} \\
\cmidrule(r){2-3} \cmidrule(r){4-5} \cmidrule(r){6-7} \cmidrule(r){8-9} \cmidrule(r){10-11}
& $\mathcal{R}^2$ $\uparrow$ & MAE $\downarrow$ & $\mathcal{R}^2$ $\uparrow$ & MAE $\downarrow$ & $\mathcal{R}^2$ $\uparrow$ & MAE $\downarrow$ & $\mathcal{R}^2$ $\uparrow$ & MAE $\downarrow$ & $\mathcal{R}^2$ $\uparrow$ & MAE $\downarrow$ \\
\midrule
LR & 0.47$\pm$0.02 & 1.32$\pm$0.02 & 0.51$\pm$0.01 & 1.29$\pm$0.01 & 0.80$\pm$0.00 & 0.72$\pm$0.02 & 0.40$\pm$0.03 & 0.88$\pm$0.03 & 0.54$\pm$0.15 & 1.05$\pm$0.26 \\
SVR & 0.67$\pm$0.02 & 1.02$\pm$0.04 & 0.77$\pm$0.01 & 1.19$\pm$0.03 & 0.71$\pm$0.01 & 0.87$\pm$0.02 & 0.44$\pm$0.06 & 0.82$\pm$0.06 & 0.65$\pm$0.13 & 0.98$\pm$0.14 \\
PR, degree 2 & 0.57$\pm$0.01 & 1.16$\pm$0.01 & 0.71$\pm$0.01 & 0.96$\pm$0.02 & 0.81$\pm$0.02 & 0.67$\pm$0.04 & 0.63$\pm$0.02 & 0.64$\pm$0.02 & 0.68$\pm$0.09 & 0.86$\pm$0.21 \\
PR, degree 3 & 0.69$\pm$0.01 & 0.97$\pm$0.02 & 0.81$\pm$0.01 & 0.77$\pm$0.02 & 0.89$\pm$0.01 & 0.51$\pm$0.02 & 0.77$\pm$0.02 & 0.48$\pm$0.02 & 0.79$\pm$0.07 & 0.68$\pm$0.20 \\
PR, degree 4 & 0.77$\pm$0.03 & 0.82$\pm$0.06 & 0.85$\pm$0.01 & 0.78$\pm$0.04 & 0.86$\pm$0.01 & 0.60$\pm$0.02 & 0.63$\pm$0.03 & 0.61$\pm$0.02 & 0.78$\pm$0.09 & 0.70$\pm$0.10 \\
PR, degree 5 & 0.73$\pm$0.03 & 0.88$\pm$0.05 & 0.83$\pm$0.02 & 0.84$\pm$0.06 & 0.84$\pm$0.02 & 0.63$\pm$0.03 & 0.57$\pm$0.05 & 0.66$\pm$0.03 & 0.74$\pm$0.11 & 0.75$\pm$0.11 \\
RF & 0.79$\pm$0.02 & 0.69$\pm$0.02 & 0.94$\pm$0.01 & 0.48$\pm$0.03 & 0.90$\pm$0.01 & 0.45$\pm$0.01 & 0.94$\pm$0.01 & 0.23$\pm$0.01 & 0.89$\pm$0.06 & 0.46$\pm$0.16 \\
GP & 0.91$\pm$0.01 & 0.50$\pm$0.02 & 0.93$\pm$0.01 & 0.54$\pm$0.03 & 0.92$\pm$0.01 & 0.40$\pm$0.02 & 0.91$\pm$0.01 & 0.30$\pm$0.01 & 0.92$\pm$0.01 & 0.43$\pm$0.09 \\
LightGBM & 0.92$\pm$0.01 & 0.44$\pm$0.03 & 0.96$\pm$0.01 & 0.38$\pm$0.02 & 0.95$\pm$0.01 & 0.32$\pm$0.02 & 0.95$\pm$0.01 & 0.20$\pm$0.01 & 0.95$\pm$0.02 & 0.34$\pm$0.09 \\
NN & \textbf{0.98$\pm$0.00} & \textbf{0.21$\pm$0.02} & \textbf{0.99$\pm$0.00} & \textbf{0.19$\pm$0.02} & \textbf{0.98$\pm$0.00} & \textbf{0.15$\pm$0.02} & \textbf{0.99$\pm$0.00} & \textbf{0.09$\pm$0.01} & \textbf{0.98$\pm$0.00} & \textbf{0.16$\pm$0.04} \\
\bottomrule
\end{tabular}
}
\caption{Surrogate models cross-validated $\mathcal{R}^2$ and MAE results. Training set $n=512$.}
\end{table}
\begin{table}[htbp]
\centering
\resizebox{\textwidth}{!}{%
\begin{tabular}{lcccccccccc}
\toprule
\multirow{2}{*}{Model} & \multicolumn{2}{c}{Axial Rotation} & \multicolumn{2}{c}{Extension} & \multicolumn{2}{c}{Flexion} & \multicolumn{2}{c}{Lateral Bending} & \multicolumn{2}{c}{\textit{Mean}} \\
\cmidrule(r){2-3} \cmidrule(r){4-5} \cmidrule(r){6-7} \cmidrule(r){8-9} \cmidrule(r){10-11}
& $\mathcal{R}^2$ $\uparrow$ & MAE $\downarrow$ & $\mathcal{R}^2$ $\uparrow$ & MAE $\downarrow$ & $\mathcal{R}^2$ $\uparrow$ & MAE $\downarrow$ & $\mathcal{R}^2$ $\uparrow$ & MAE $\downarrow$ & $\mathcal{R}^2$ $\uparrow$ & MAE $\downarrow$ \\
\midrule
LR & 0.48$\pm$0.00 & 1.30$\pm$0.01 & 0.50$\pm$0.00 & 1.30$\pm$0.01 & 0.80$\pm$0.00 & 0.72$\pm$0.00 & 0.41$\pm$0.02 & 0.88$\pm$0.01 & 0.55$\pm$0.15 & 1.05$\pm$0.26 \\
SVR & 0.60$\pm$0.01 & 1.13$\pm$0.01 & 0.75$\pm$0.00 & 0.99$\pm$0.01 & 0.62$\pm$0.01 & 0.98$\pm$0.02 & 0.09$\pm$0.06 & 1.06$\pm$0.04 & 0.51$\pm$0.25 & 1.04$\pm$0.06 \\
PR, degree 2 & 0.57$\pm$0.01 & 1.14$\pm$0.01 & 0.73$\pm$0.01 & 0.93$\pm$0.00 & 0.84$\pm$0.00 & 0.62$\pm$0.01 & 0.67$\pm$0.02 & 0.59$\pm$0.02 & 0.70$\pm$0.10 & 0.82$\pm$0.23 \\
PR, degree 3 & 0.70$\pm$0.01 & 0.94$\pm$0.02 & 0.84$\pm$0.01 & 0.71$\pm$0.03 & 0.92$\pm$0.01 & 0.43$\pm$0.02 & 0.84$\pm$0.02 & 0.37$\pm$0.02 & 0.82$\pm$0.08 & 0.61$\pm$0.23 \\
PR, degree 4 & 0.85$\pm$0.01 & 0.64$\pm$0.02 & 0.91$\pm$0.01 & 0.57$\pm$0.03 & 0.93$\pm$0.01 & 0.40$\pm$0.01 & 0.87$\pm$0.02 & 0.38$\pm$0.04 & 0.89$\pm$0.03 & 0.50$\pm$0.11 \\
PR, degree 5 & 0.83$\pm$0.01 & 0.67$\pm$0.02 & 0.91$\pm$0.01 & 0.61$\pm$0.03 & 0.91$\pm$0.01 & 0.46$\pm$0.01 & 0.84$\pm$0.02 & 0.42$\pm$0.04 & 0.87$\pm$0.04 & 0.54$\pm$0.10 \\
RF & 0.85$\pm$0.03 & 0.54$\pm$0.03 & 0.95$\pm$0.01 & 0.45$\pm$0.03 & 0.90$\pm$0.02 & 0.43$\pm$0.04 & 0.96$\pm$0.00 & 0.20$\pm$0.01 & 0.91$\pm$0.04 & 0.40$\pm$0.12 \\
GP & 0.94$\pm$0.00 & 0.40$\pm$0.02 & 0.96$\pm$0.01 & 0.43$\pm$0.03 & 0.95$\pm$0.00 & 0.31$\pm$0.01 & 0.96$\pm$0.01 & 0.19$\pm$0.02 & 0.95$\pm$0.01 & 0.33$\pm$0.09 \\
LightGBM & 0.95$\pm$0.01 & 0.34$\pm$0.03 & 0.97$\pm$0.00 & 0.33$\pm$0.02 & 0.96$\pm$0.01 & 0.29$\pm$0.02 & 0.97$\pm$0.01 & 0.16$\pm$0.01 & 0.96$\pm$0.01 & 0.28$\pm$0.07 \\
NN & \textbf{0.99$\pm$0.00} & \textbf{0.13$\pm$0.01} & \textbf{1.00$\pm$0.00} & \textbf{0.11$\pm$0.01} & \textbf{0.99$\pm$0.00} & \textbf{0.10$\pm$0.01} & \textbf{1.00$\pm$0.00} & \textbf{0.06$\pm$0.00} & \textbf{0.99$\pm$0.00} & \textbf{0.10$\pm$0.03} \\
\bottomrule
\end{tabular}
}
\caption{Surrogate models cross-validated $\mathcal{R}^2$ and MAE results. Training set $n=1024$.}
\end{table}

\begin{table}[H]
\centering
\resizebox{\textwidth}{!}{%
\begin{tabular}{lcccccccc}
\toprule
\multirow{2}{*}{Moments (Nm)} & \multicolumn{2}{c}{Random Forest (RF)} & \multicolumn{2}{c}{Gaussian Process (GP)} & \multicolumn{2}{c}{LightGBM} & \multicolumn{2}{c}{Neural Network (NN)} \\
\cmidrule(r){2-3} \cmidrule(r){4-5} \cmidrule(r){6-7} \cmidrule(r){8-9}
 & $\overline{\mathcal{R}^2}$ $\uparrow$ & MAE $\downarrow$ & $\overline{\mathcal{R}^2}$ $\uparrow$ & MAE $\downarrow$ & $\overline{\mathcal{R}^2}$ $\uparrow$ & MAE $\downarrow$ & $\overline{\mathcal{R}^2}$ $\uparrow$ & MAE $\downarrow$ \\
\midrule
1, 2, 3, 4, 5 & 0.92$\pm$0.01 & 0.01$\pm$0.00 & 0.95$\pm$0.00 & 0.01$\pm$0.00 & 0.96$\pm$0.00 & 0.01$\pm$0.00 & \textbf{0.99$\pm$0.00} & \textbf{0.00$\pm$0.00} \\
0.5 & -2.10$\pm$0.12 & 0.03$\pm$0.00 & \textbf{0.85$\pm$0.01} & \textbf{0.00$\pm$0.00} & -1.85$\pm$0.18 & 0.03$\pm$0.00 & 0.43$\pm$0.01 & 0.01$\pm$0.00 \\
1.5, 2.5, 3.5, 4.5 & 0.87$\pm$0.01 & 0.02$\pm$0.00 & 0.95$\pm$0.00 & 0.01$\pm$0.00 & 0.91$\pm$0.01 & 0.02$\pm$0.00 & \textbf{0.99$\pm$0.00} & \textbf{0.00$\pm$0.00} \\
$> 5$ & 0.69$\pm$0.02 & 0.05$\pm$0.00 & 0.82$\pm$0.01 & 0.04$\pm$0.00 & 0.73$\pm$0.01 & 0.05$\pm$0.00 & \textbf{0.95$\pm$0.01} & \textbf{0.02$\pm$0.00} \\
\bottomrule
\end{tabular}
}
\caption{Comparison of interpolation and extrapolation ability of the top performing surrogate models. The results are the mean and standard deviation across the folds.}
\label{table:interextra}
\end{table}

\begin{table}[H]
\centering
\resizebox{\textwidth}{!}{%
\begin{tabular}{llccccccccc}
\toprule
\multirow{2}{*}{Sample} & \multirow{2}{*}{} &  \multirow{2}{*}{Method} & \multicolumn{2}{c}{Axial Rotation} & \multicolumn{2}{c}{Extension} & \multicolumn{2}{c}{Flexion} & \multicolumn{2}{c}{Lateral Bending} \\
\cmidrule(r){4-5} \cmidrule(r){6-7} \cmidrule(r){8-9} \cmidrule(r){10-11}
& & & $\mathcal{R}^2$ $\uparrow$ & MAE $\downarrow$ & $\mathcal{R}^2$ $\uparrow$ & MAE $\downarrow$ & $\mathcal{R}^2$ $\uparrow$ & MAE $\downarrow$ & $\mathcal{R}^2$ $\uparrow$ & MAE $\downarrow$ \\
\midrule
\multirow{10}{*}{\citet{nicolini2022effects}} & \multirow{2}{*}{1} & PGD w/ NN (Ours) & \textbf{0.94} & \textbf{0.43} & \textbf{0.98} & \textbf{0.29} & \textbf{0.98} & \textbf{0.28} & \textbf{0.99} & \textbf{0.11}\\
 &  & GA w/ NN & 0.94 & 0.46 & 0.93 & 0.54 & 0.95 & 0.44 & 0.97 & 0.23 \\
\cmidrule(r){2-11}
& \multirow{2}{*}{2} & PGD w/ NN (Ours) & 0.85 & 0.13 & \textbf{0.99} & \textbf{0.03} & \textbf{0.99} & \textbf{0.08} & 0.99 & 0.06 \\
 &  & GA w/ NN & \textbf{0.99} & \textbf{0.03} & 0.97 & 0.15 & 0.95 & 0.18 & \textbf{0.99} & \textbf{0.02} \\
\cmidrule(r){2-11}
& \multirow{2}{*}{3} & PGD w/ NN (Ours) & 0.70 & 1.05 & 0.60 & 2.15 & \textbf{0.97} & \textbf{0.57} & \textbf{0.91} & \textbf{0.66} \\
 &  & GA w/ NN & \textbf{0.99} & \textbf{0.14} & \textbf{0.98} & \textbf{0.47} & 0.95 & 0.70 & 0.80 & 1.17 \\
\cmidrule(r){2-11}
& \multirow{2}{*}{4} & PGD w/ NN (Ours) & 0.87 & 0.46 & \textbf{0.99} & \textbf{0.16} & \textbf{0.95} & \textbf{0.37} & 0.82 & 0.42 \\
 &  & GA w/ NN & \textbf{0.93} & \textbf{0.31} & 0.94 & 0.35 & 0.88 & 0.59 & \textbf{0.95} & \textbf{0.21} \\
\cmidrule(r){2-11}
& \multirow{2}{*}{5} & PGD w/ NN (Ours) & \textbf{0.96} & \textbf{0.14} & \textbf{0.99} & \textbf{0.09} & \textbf{0.99} & \textbf{0.05} & 0.81 & 0.52 \\
 &  & GA w/ NN & 0.96 & 0.15 & 0.99 & 0.16 & 0.93 & 0.36  & \textbf{0.86} & \textbf{0.42} \\
\midrule
\multirow{2}{*}{\citet{heuer2007stepwise}} & \multirow{2}{*}{Mean} & PGD w/ NN (Ours) & \textbf{0.98} & \textbf{0.19} & \textbf{0.94} & \textbf{0.82} & \textbf{0.97} & \textbf{0.45} & \textbf{0.99} & \textbf{0.1}\\
 &  & GA w/ NN & 0.94 & 0.42 & 0.92 & 0.99 & 0.95  & 0.56 & 0.98 & 0.24 \\
\bottomrule
\end{tabular}
}
\caption{Breakdown across load cases of calibration methods on experimental specimens.}
\end{table}

\begin{figure}[H]
    \centering
    \includegraphics[width=\textwidth]{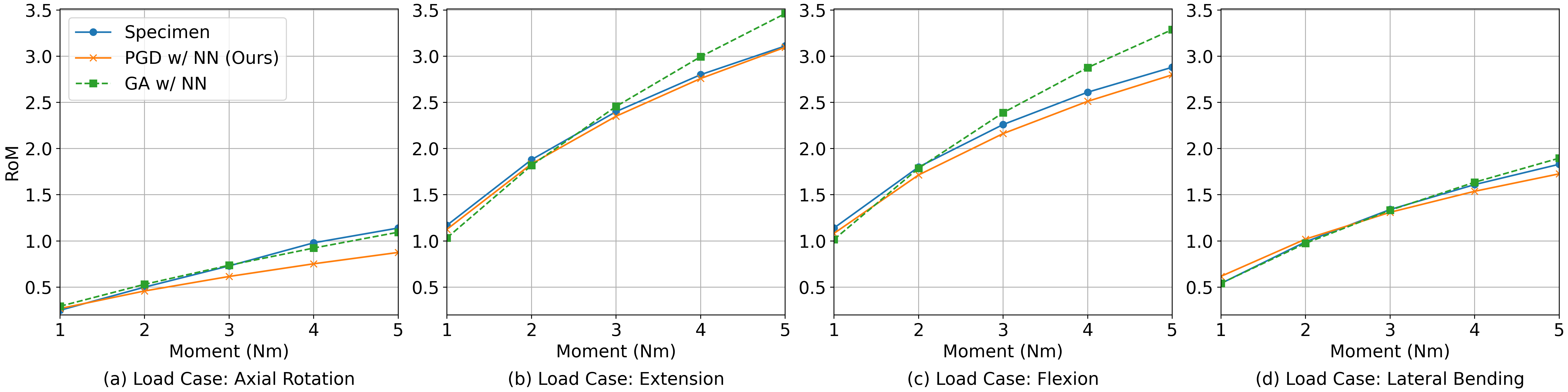}
    \caption{Calibration of \citet{nicolini2022effects} sample 2}
\end{figure}

\begin{figure}[H]
    \centering
    \includegraphics[width=\textwidth]{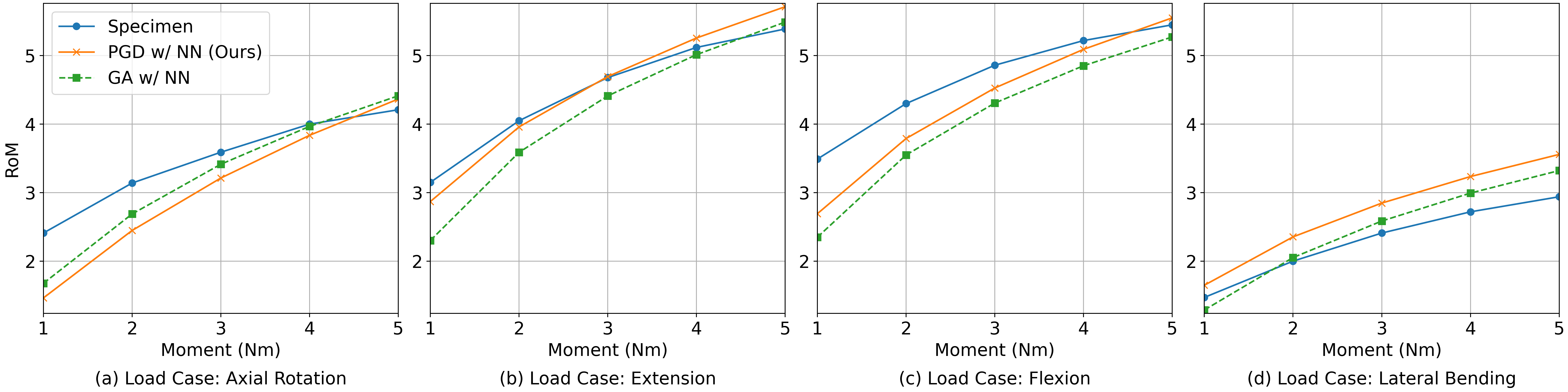}
    \caption{Calibration of \citet{nicolini2022effects} sample 4}
\end{figure}

\begin{figure}[H]
    \centering
    \includegraphics[width=\textwidth]{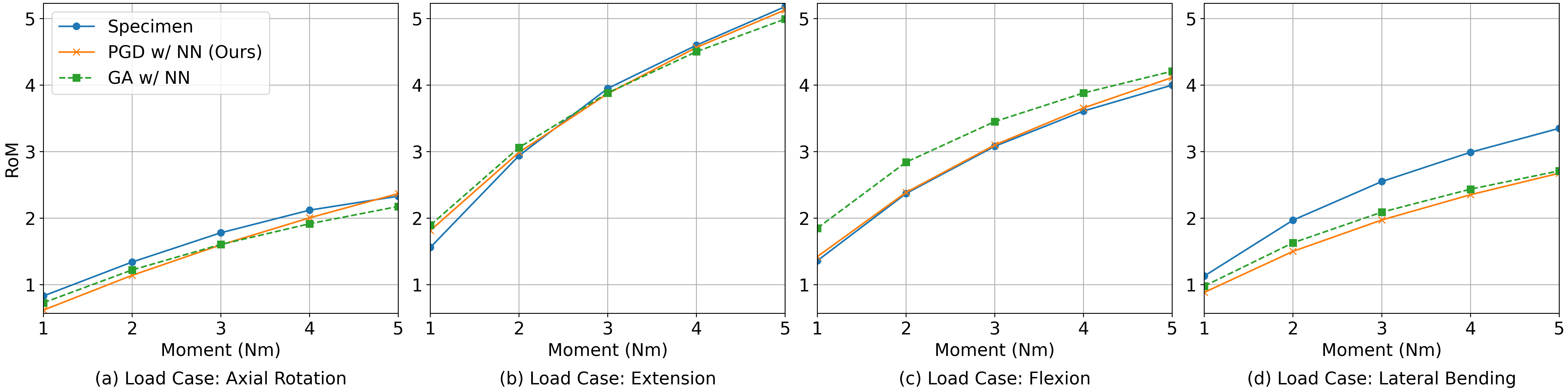}
    \caption{Calibration of \citet{nicolini2022effects} sample 5}
\end{figure}

\begin{figure}[H]
    \centering
    \includegraphics[width=\textwidth]{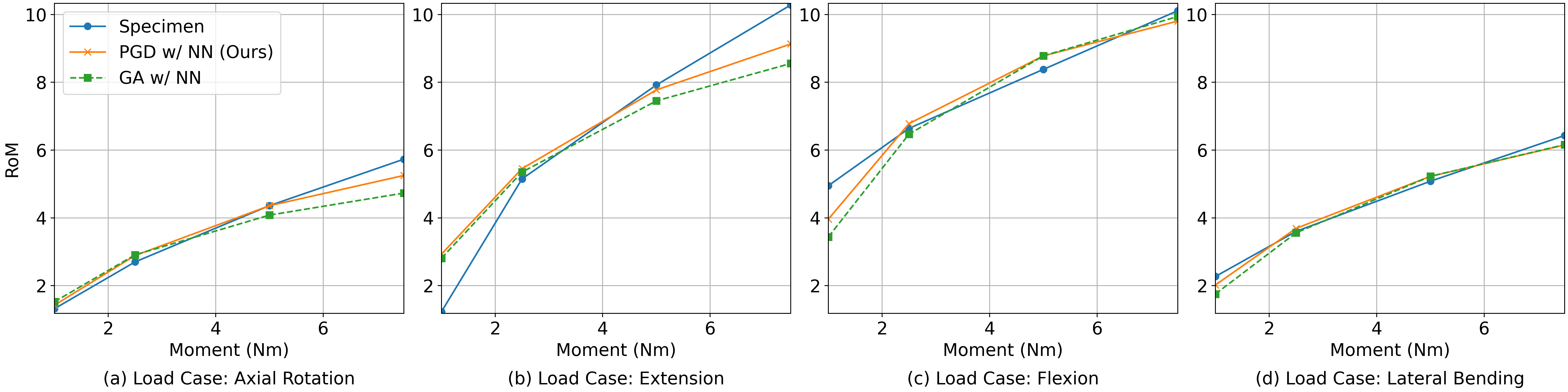}
    \caption{Calibration of \citet{heuer2007stepwise} samples mean}
\end{figure}
        
\begin{table}[H]
\centering
\resizebox{\textwidth}{!}{%
\begin{tabular}{llccccccccccccc}
\toprule
Sample &  & $C_{10n}$ &  $C_{01n}$ &  $C_{10a}$ &  $k_1$ &  $k_2$ &    $\kappa$ &    $k_{1c}$ &    $k_{2c}$ &     $k_{1r}$ &     $k_{2r}$ &  $\alpha$ &  $\alpha_c$ &  $\alpha_r$ \\
\midrule
  \multirow{5}{*}{\citet{nicolini2022effects}} & 1 &    0.043 &    0.007 &    0.060 &  50.349 & 199.739 & 0.268 & -0.200 & -0.166 & -0.041 & -0.019 &   24.054 &        0.000 &       0.041 \\
  & 2 &    0.039 &    0.048 &    0.072 &  38.859 & 190.633 & 0.003 & -0.054 & -0.133 & -0.091 & -0.026 &   27.157 &        0.255 &       0.156 \\
  & 3 &    0.117 &    0.019 &    0.066 &   1.556 & 172.572 & 0.116 & -0.107 & -0.025 & -0.133 & -0.122 &   26.626 &        0.015 &       0.116 \\
  & 4 &    0.085 &    0.037 &    0.067 &  93.089 & 181.426 & 0.267 & -0.130 & -0.201 & -0.187 & -0.021 &   22.690 &        0.064 &       0.181 \\
  & 5 &    0.211 &    0.009 &    0.216 & 104.187 &  97.086 & 0.233 & -0.166 & -0.001 & -0.189 & -0.159 &   26.384 &        0.231 &       0.023 \\
 \citet{heuer2007stepwise} & Mean &  0.062 &    0.028 &    0.086 &  11.800 &  66.969 & 0.162 & -0.197 & -0.006 & -0.023 & -0.186 &   38.925 &        0.017 &       0.009 \\
\bottomrule
\end{tabular}
}
\caption{Result calibrated configurations for experimental data with our method.}
\end{table}

\section{Running time analysis}\label{appendix:running_time}

PGD w/ NN requires an initial dataset generation step, where 1024 samples are created, each with four FE simulations (one per load case), resulting in 4096 total simulations. Each simulation takes approximately 208 seconds, leading to a total dataset generation time of 236.9 hours or 9.87 days. The model training time is negligible at 1.03 minutes, and once trained, calibration takes only 2.58 seconds, irrespective for the number of specimens.

\begin{figure}[htbp]
    \centering
    \includegraphics[width=0.8\textwidth]{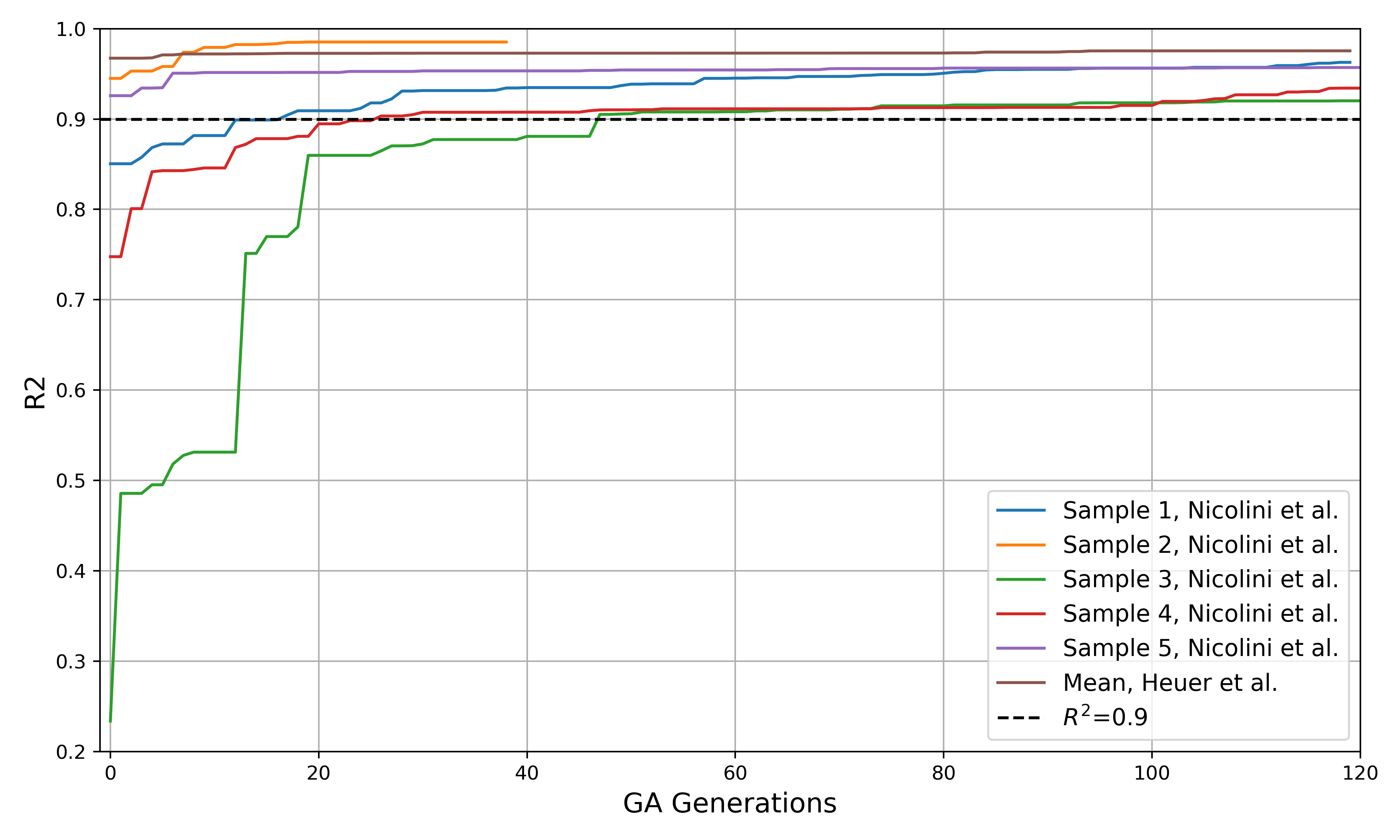}
    \caption{\textbf{GA w/NN evolution of $\overline{\mathcal{R}^2}$ values for experimental specimens.} 3rd sample by \citet{nicolini2022effects} requires more than double the amount of generations to converge to a score higher than 0.9.}
    \label{fig:ga_convergence}
\end{figure}

For a traditional GA setup, using the FE simulations and not the surrogate predictions, we derive worst-case and average-case estimates following the convergence plot in Fig.~\ref{fig:ga_convergence}. We assume that 16 (average-case) or 47 (worst-case) generations would be required for the algorithm to converge to an $\overline{\mathcal{R}^2}$ higher than 0.9. These result in $20 + (1 \times 19) + (14 \times 14) = 235$ or $20 + (4 \times 19) + (42 \times 14) = 684$ individuals, with each individual requiring four simulations as before. Here simulations are longer than for the NN, taking 250 seconds (as some specimens involve higher moments), resulting in 2.7 (average-case) or 7.9 (worst-case) days per specimen.

\clearpage

\section{List of abbreviations}

\begin{itemize}
    \item \textbf{FE} - Finite Element
    \item \textbf{GA} - Genetic Algorithm
    \item \textbf{GP} - Gaussian Process
    \item \textbf{IVD} - Intervertebral Disc
    \item \textbf{LHS} - Latin hypercube sampling
    \item \textbf{LR} - Linear Regression
    \item \textbf{MAE} - Mean Absolute Error
    \item \textbf{ML} - Machine Learning
    \item \textbf{MPa} - Megapascal
    \item \textbf{Nm} - Newton meter
    \item \textbf{NN} - Neural Network
    \item \textbf{PGD} - Projected Gradient Descent
    \item \textbf{PR} - Polynomial Regression
    \item \textbf{RF} - Random Forest
    \item \textbf{RoM} - Range of Motion
    \item \textbf{SVR} - Support Vector Machine for Regression
\end{itemize}

\clearpage

\bibliographystyle{elsarticle-num-names} 
\bibliography{cas-refs}

\end{document}